\RequirePackage[T1,T2A]{fontenc}

\documentclass[fleqn,12pt]{wlscirep}

\usepackage[utf8]{inputenc}

\AtBeginDocument{\fontencoding{T1}\selectfont}

\usepackage{geometry}
\usepackage{graphicx}
\usepackage{subcaption}
\usepackage{float}
\usepackage{dblfloatfix}

\usepackage{booktabs}
\usepackage{longtable}
\usepackage{array}
\usepackage{adjustbox}
\usepackage{makecell}
\usepackage{colortbl} 
\usepackage{longtable}

\usepackage{pdflscape}

\usepackage[section]{placeins}

\usepackage{tikz}
\usetikzlibrary{arrows.meta, positioning, shapes.geometric, shapes.misc, fit}

\usepackage{xurl}
\usepackage{hyperref}

\newcommand{\rutt}[1]{{\begingroup
  \fontencoding{T2A}\ttfamily\selectfont #1%
\endgroup}}

\usetikzlibrary{arrows.meta,positioning,shapes.geometric,fit,calc}

\usetikzlibrary{shapes, arrows.meta, positioning, shapes.geometric, arrows}
\title{Linked Multi-Modal Data on Russian Domestic and Foreign Policy Speeches} 

\author[1]{Daria Blinova}
\author[2]{Gayathri Emuru}
\author[2]{Rakesh Emuru}
\author[2]{Kushagradheer Shridheer Srivastava}
\author[3]{Mina Rulis}
\author[4,]{Sunita Chandrasekaran}
\author[1,*]{Benjamin E. Bagozzi}
\affil[1]{University of Delaware, Department of Political Science \& International Relations, Newark, DE, USA}
\affil[2]{University of Delaware, Masters of Science in Data Science Program, Newark, DE, USA}
\affil[3]{University of Pennsylvania, Department of Political Science, Philadelphia, PA, USA}
\affil[4]{University of Delaware, Department of Computer \& Information Sciences, Newark, DE, USA}

\affil[*]{corresponding author: Benjamin E. Bagozzi (bagozzib@udel.edu)}

\begin{abstract}
This paper introduces a dataset of interlinked multimodal political communications from the Russian government, addressing  persistent deficiencies in the availability of social text- and image-based data for authoritarian politics contexts. The dataset comprises two large corpora of official speeches delivered by senior actors within the Kremlin and the Russian Ministry of Foreign Affairs over multiple decades. For each speech, we provide Russian- and English-language texts, associated images and captions where available, and harmonized metadata including (e.g.) dates, speakers, (geo)locations, and official government content tags. Unique identifiers link images to speeches and align Russian and English versions of the same communication texts. We further augment these linked datasets with validated topical annotations for both speech texts and speech images, which are generated via transformer-based multimodal topic modeling and refined by a Russian politics expert. The resulting data resources support multimodal, multilingual, temporal, and/or spatial analyses of (authoritarian) political communication and offer a valuable testbed for social science research and large language model (LLM) applications in political domains.
\end{abstract}
\begin{document}

\flushbottom
\maketitle

\thispagestyle{empty}

\section*{Background \& Summary}

Over the past decade, social scientists have increasingly turned to text-as-data and, more recently, image-as-data approaches to study political, social, and economic phenomena at scale. Advances in natural language processing (NLP), computer vision, and representation learning have enabled researchers to analyze vast corpora of speeches, news articles, social media posts, and visual content that were previously inaccessible to systematic empirical analysis. In political science, this shift has reshaped the study of elite behavior, public opinion, and international relations, with influential contributions appearing in \textit{Political Analysis}, \textit{American Political Science Review}, \textit{American Journal of Political Science}, and the \textit{Annual Review of Political Science} \cite{GrimmerStewart2013,wilkerson2017large,Mueller2018,benoit2019measuring,Torres2024,Arjun2025}. Parallel developments are evident across sociology \cite{steinert2019future,bonikowski2022ends}, communications \cite{Casas2022,birkenmaier2024search}, management and organizational research \cite{li2024computer,shahgholian2023big}, and psychology \cite{paulus2016positive,bittermann2024natural}, underscoring trends towards disciplinary convergence around computational approaches to social inquiry. Concurrently, these methods have become central to computer science and data science research itself, where socially grounded text and image corpora are increasingly used to evaluate models, study bias, and develop multimodal learning techniques \cite{Mahajan_2018_ECCV,radford2021learning,xu2023mplug}.

Despite this progress, scholars still lack comprehensive and systematically linked social text- and image-based datasets. This gap is particularly acute for datasets that combine linked multimodal content---that is, text and images associated with the same political communication---and for datasets that include multilingual text, which have each been shown to be critical for understanding political messaging, framing, and signaling \cite{KressVanLeeuwen2001,BaltrusaitisEtAl2019,Pineda2023,LiuShao2024}. The absence of such data is especially consequential in authoritarian settings, where conventional sources of quantitative information---such as economic statistics, public opinion surveys, or administrative records---are often unavailable, selectively released, or strategically manipulated \cite{HollyerEtAl2011,Wallace_2016,RozenasEtal2019}. Existing exceptions that rely on either text or images alone---such as analyses of authoritarian rhetoric in online media or speech or studies of state propaganda imagery---have nonetheless demonstrated substantial analytic value, illuminating patterns of elite signaling, policy priorities, and regime legitimation strategies \cite{carroll2017image,Roberts2018,RozenasEtal2019,mochtak2025chasing,la2025text,Zhong16102025}. In short, these latter studies suggest that richer, multimodal, and multilingual data could further deepen our understandings of authoritarian politics.

Motivated by this potential, we collect and curate a novel set of interlinked multimodal datasets combining text, images, and metadata from two distinct collections of Russian government speeches. Our overarching data resource includes linked Russian- and English-language versions of official speech texts, associated speech images, and contextual metadata for speeches delivered by high-level political actors (most commonly, the Russian president) under the auspices of the Kremlin during the period 31 December 1999 to 20 September 2025, as well as speeches delivered by high-level political actors (most commonly, the Minister of Foreign Affairs of the Russian Federation) under the auspices of the Russian Ministry of Foreign Affairs from 18 March 2004 to 7 October 2025. Across these two corpora, the data encompass 15,610 total English-language speeches and 19,396 total Russian-language speeches, along with associated sets of 42,782, and 49,277 images, providing rich multi-modal content over considerable time spans covering key periods in contemporary Russian domestic politics and international relations.

In total, we release cleaned Russian- and English-language texts, all associated images, and harmonized metadata for both these Kremlin and Ministry of Foreign Affairs speech collections. Each dataset includes unique identifiers linking images to specific speeches, as well as identifiers linking Russian and English versions of the same speech. The Kremlin and Russian Ministry of Foreign Affairs translate and provide these separate Russian and English translated versions of our collected speeches themselves. However, as discussed further below, these parallel English and Russian speech versions are not necessarily identical. This ensures that our collected and linked data will uniquely allow future scholars to study variation in official Russian government decisions concerning which content to include or omit within each language version of a given speech, and the potential causes and effects of such decisions. Where available, our extracted metadata for each speech corpus also include the date of each speech, official indexing tags assigned by the Russian government (e.g., for theme, region, and speaker), speech titles and summaries, image captions, the named speech location, speaker names, and our own extracted and validated speech geolocations---alongside additional (meta)data to aid researchers.

Beyond these textual, image, and metadata attributes, we also enrich each corpora with our own substantively meaningful topic annotations that we develop via a human-in-the-loop framework. Using transformer-based multimodal topic modeling---specifically BERTopic \cite{Grootendorst2022}---we estimate latent topics separately for each (language-specific) text corpus and its associated image data. A Russian politics subject-matter expert then labels the core topics and groups them into higher-level thematic categories. We validate these topics through extensive comparisons with official Kremlin-provided thematic labels (where available) and secondary qualitative reviews by political science subject matter experts. Together our topic annotations provide transparently-extracted and near-complete coverage of all speeches in our Kremlin and Ministry of Foreign Affairs corpus, in contrast to the Russian government's own thematic topic labels of a limited portion of Kremlin-specific speeches.

Altogether, resulting data offer a wide range of applications for social scientists, computer scientists, and data scientists alike. For social scientists, the linked texts, images, and topic labels enable systematic analyses of Russian domestic and foreign policy priorities over time and potentially space, extending prior work that has relied almost exclusively on single-language, text-only corpora from a single ministry or executive actor \cite{Mölder2023,Blinova2025,Yavuz2025}. Together these data  accordingly allow social science researchers to study the political origins and/or effects of (i) divergences in speech (text and or image) content across Russia's Kremlin and Ministry of Foreign Affairs and (ii) divergences in the English versus Russian language versions of each speech released by these respective political units. These potential comparisons speak directly to a growing literature on authoritarian signaling and audience differentiation, including work on China showing how regimes tailor messages to foreign and domestic publics \cite{Weiss_2013,WeissDafoe2019,Dai2022, Liu2023, Mochtak2021}.

Likewise, many data science, AI, and computer science researchers rely on large test beds of annotated texts (and images) for developing or validating new machine learning or AI methods. Our expert-labeled and validated topic variables, and linked English- and Russian-datasets more generally, provide ready-to-use inputs for such tasks, especially when these tasks intend to explore qualities of Russian-language text and/or multi-lingual or multimodal content. And finally for government officials---as well as researchers interested in political forecasting---the text, image, and topic information in our data can be used to develop inputs for regression analyses, time-series models, or forecasting and early-warning systems related to conflict, foreign policy behavior, and state stability \cite{o2002anticipating,Blair2020, d2020conflict,Mueller2018}.

In what follows, we first review the project scope and documents, followed by an overview of our webscraping, data cleaning, and measurement strategies for our texts, images, and metadata. Next, we provide details on the structure of our dataset and data records. Lastly, we discuss our validation exercises and conclude with the usage notes and details on data and code availability.

\section*{Methods}

\subsection*{Project Scope}

In this section, we describe the sets of text and image corpora that we extract from the Kremlin (\texttt{kremlin.ru}) and the Russian Ministry of Foreign Affairs (\texttt{mid.ru}) official websites. The Kremlin is the official representation of the Russian President, who is the executive head of the Russian state. The Ministry of Foreign Affairs is a federal executive authority implementing foreign policy and operating under the jurisdiction of the Russian President. Both of these Russian executive branch institutions serve complementary roles in shaping the direction of the Russian state’s internal and/or external affairs. At the same time, they serve as a source of important political information, allowing outside observers to assess the dynamics of the state’s official rhetoric. Since both the Kremlin and the Ministry of Foreign Affairs operate their own official websites, they separately archive speeches, press releases, interviews, and other relevant content involving texts and images. Though these speeches and images are also at times disseminated in other mediums, such as through television media, these two governmental websites are unique in the extent to which they archive these data in a systematic manner over extended periods. Our extraction and released materials focus on the textual content of each item and any associated still images; we do not extract, store, or analyze video content. Below, we first describe the English and Russian Kremlin corpora (and accompanying images). We then turn to describe the English and Russian versions of the Ministry of Foreign Affairs corpora (and its images).

\subsection{Kremlin Texts and Images}

The Kremlin text and image corpora are extracted from the official Kremlin website (\texttt{kremlin.ru}). This website archives transcripts of all Kremlin speeches (and their visuals) given on different occasions from 31 December 1999 onward, which for the purposes of our data collection covers up to and including 20 September 2025. The Kremlin website stores speech transcripts both in Russian and English, alongside images that reflect the speech setting corresponding to the speech content. As discussed in more detail above and below, we extract these English and Russian speech texts and their visuals in separate datasets and provide additional metadata accompanying these extracted texts and images.  

Given the nature of the Kremlin and its official website, a majority of the Russian and English speeches on this website are given by the President of Russia. During the date range of our dataset, Russia had only two presidents: Vladimir Putin (2000-2008; 2012-Present) and Dmitry Medvedev (2008-2012). These actors represent a majority of the speakers recorded across the speeches associated with the Kremlin and its official website. Yet in addition to presidential speeches, the Kremlin's available transcripts also include a smaller share of speeches from domestic or international political representatives with whom these two presidents interacted. In such cases, these transcripts typically include speech content from the Russian president as part of a collaboration (e.g.., meeting or joint speech) with such leaders.

As noted previously, the occasions on which these public speeches are given vary, as does the format of the speech (and accompanying image) itself and its thematic focus. Such variation spans interviews or official communication between the president and domestic or international leaders to nationwide announcements and transcriptions of international forum performances. At the same time, the events where such speeches are given are equally diverse and include bilateral meetings, multilateral arrangements, domestic events, presidential greetings and salutations, addresses for national holidays, and others.

Each archived speech on the Kremlin's website includes a speech title, the speech text itself, a date (down to calendar day) and time of day, a location, and (for more recent speeches) a set of thematic tags assigned by the Kremlin itself. In some cases, as noted earlier, speeches can also contain images with captions. As discussed in more detail below, much of the above content is incorporated as metadata alongside the main text and image data for our Kremlin corpora. Finally, we can note that while some prior research has analyzed the English versions of the Kremlin's speech text transcripts \cite{Blinova2025, Mölder2023}, no work to our knowledge has extracted a comprehensive set of both Russian and English versions of these speech text transcripts, nor of the accompanying images.

\subsection{The Ministry of Foreign Affairs of the Russian Federation Texts and Images}

The Russian Ministry of Foreign Affairs (hereafter abbreviated as MID, given the Russian name of the ministry, \textit{Ministerstvo Inostrannyh Del}) represents a distinct source of official Russian government rhetoric. Relative to the earlier Kremlin discussion, the MID more narrowly administers Russian foreign policy priorities rather than international and domestic Russian policy concerns. The official website of the MID (\texttt{mid.ru}), which we used to extract our MID corpora and visuals, collects various Ministry speeches and media made by representatives of the agency. Our primary speeches and images of interest from this website primarily relate to the statements and speeches made by the Minister of Foreign Affairs, Sergey Lavrov, who assumed office in March 2004 and served until the time of writing. As in the case of the Kremlin website, the MID archives these texts in both Russian and English with corresponding metadata. Since Lavrov’s speech records coincide with his presence in office, the date range for our datasets runs from 18 March 2004 to 7 October 2025.

Like the Kremlin corpora, the nature of the textual speeches (and their associated images) that are made by the Russian Foreign Minister and at times other parties that he engages with during these speeches is remarkably diverse. To this end, we can note that the MID's speeches encompass settings related to press conferences that the Minister has held both within and outside the country, interviews that the mass media ask for as a result of summits or bilateral meetings, including personal interviews such as with Tucker Carlson and other prominent commentators, as well as occasional meetings within Russia's consulates and other diplomatic exchanges. 

Similar to the Kremlin speeches discussed further above, all MID speeches are systematically formatted and consistently archived in both Russian and English on the MID's official website. For each speech therein, this archived content includes a speech title, the speech text itself, any corresponding images, as well as information on the speech's recorded date, time, and location. That being said, and in comparison to the Kremlin speeches discussed earlier, this particular set of archived speeches does not contain any ministry-assigned thematic tags. Moreover, location information for the MID speeches is not always clearly stored as meta-data and hence is not as easily extractible as is the case for our Kremlin speeches. These caveats aside, the MID speech (text and image) data and available metadata altogether represent the main focus of our extraction efforts for this particular ministry as discussed further below. To the best of our knowledge, no research has extracted or considered these particular speech data across both their English and Russian language text content, nor with regards to their associated images.

\subsection*{Webscraping} 

Our first objective was to webscrape the websites discussed above in order to collected all inputs needed for the construction of four parallel corpora of official speeches and their associated images, spanning the Kremlin (Russian and English) and the Russian Ministry of Foreign Affairs (MID; Russian and English). For each source-language pair this requires that we systematically collect the page markup (HTML) for every speech, extract core text fields (identifier, URL, title, date, and main body), enumerate and download all images associated with that page, and write an analysis-ready CSV that references a per–speech image directory. We now describe these steps in further detail. Throughout this discussion, we refer to the downloaded page content as ``HTML files'' and to the tabular outputs as ``CSV files'' to keep the terminology precise and consistent.

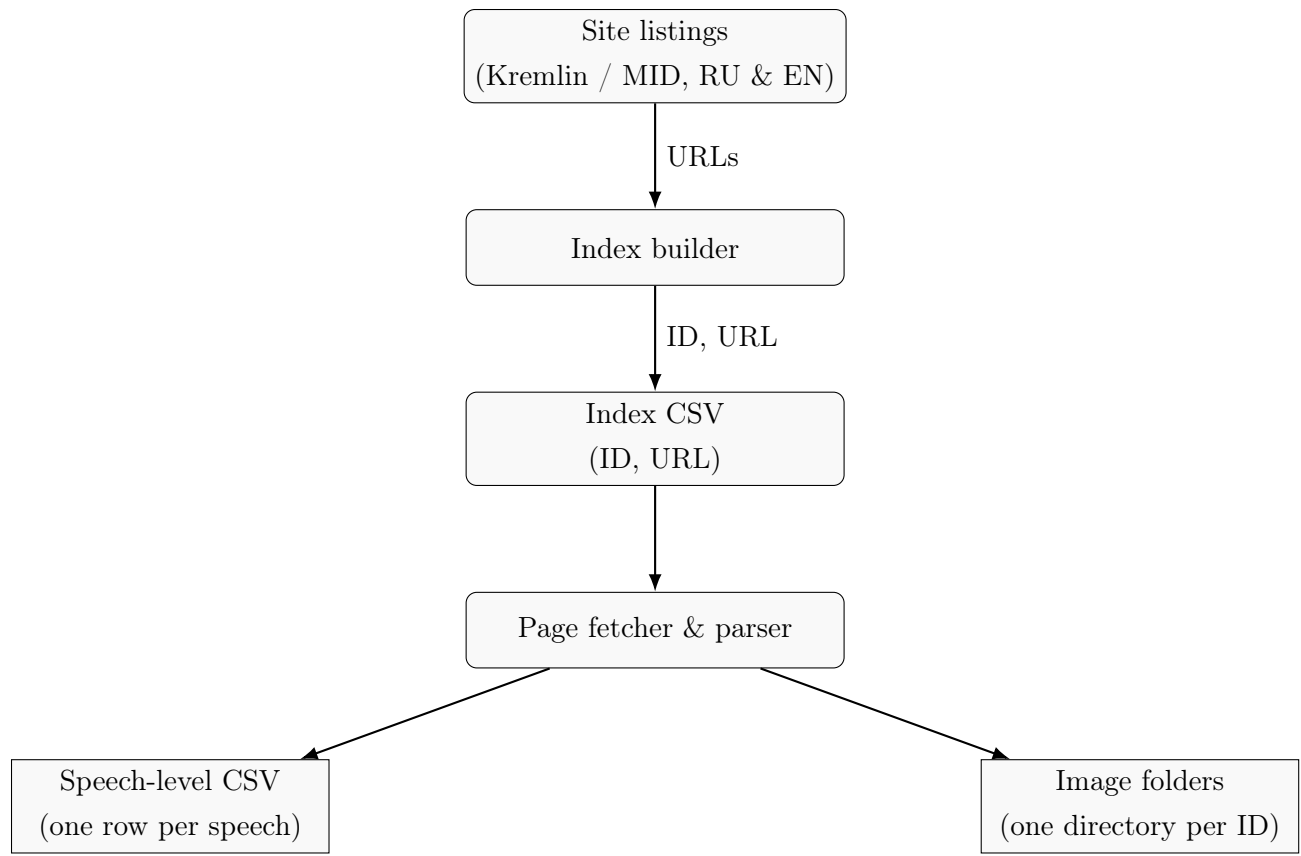
\begin{figure}[H]
    \centering
    \begin{tikzpicture}[
        font=\small,
        >=Latex,
        node distance=1.4cm,
        block/.style={
            rectangle,draw,rounded corners,
            fill=gray!5,
            align=center,
            minimum width=5.0cm,
            minimum height=1.0cm
        },
        outblock/.style={
            rectangle,draw,
            fill=gray!5,
            align=center,
            minimum width=4.2cm,
            minimum height=0.9cm
        },
        arrow/.style={->,thick}
    ]

    \node[block] (listings) {Site listings\\(Kremlin / MID, RU \& EN)};
    \node[block, below=of listings] (indexer) {Index builder};
    \node[block, below=of indexer] (indexcsv) {Index CSV\\(ID, URL)};
    \node[block, below=of indexcsv] (fetcher) {Page fetcher \& parser};

    \node[outblock, below left=1.2cm and 1.8cm of fetcher]
        (speechcsv) {Speech-level CSV\\(one row per speech)};
    \node[outblock, below right=1.2cm and 1.8cm of fetcher]
        (images) {Image folders\\(one directory per ID)};

    \draw[arrow] (listings) -- node[right]{URLs} (indexer);
    \draw[arrow] (indexer) -- node[right]{ID, URL} (indexcsv);
    \draw[arrow] (indexcsv) -- (fetcher);
    \draw[arrow] (fetcher) -- (speechcsv);
    \draw[arrow] (fetcher) -- (images);

    \end{tikzpicture}
    \caption{Two-stage webscraping workflow. For each source (Kremlin, MID) and language (Russian, English), an index builder first traverses the site listings and writes an index CSV of speech IDs and URLs. A page fetcher-parser then consumes this index, downloads each page, extracts structured text and metadata into a speech-level CSV, and saves all associated images into per-ID folders.}
    \label{fig:webscrape-pipeline}
\end{figure}

From an implementation perspective, all webscraping is carried out in \texttt{Python 3}~\cite{python3} using the \texttt{requests} ~\cite{requests2025} library for HTTP retrieval, \texttt{BeautifulSoup}~\cite{richardson_bs4} for HTML parsing, and \texttt{pandas}~\cite{mckinney2010pandas} and the Python standard library (\texttt{csv}, \texttt{pathlib}, \texttt{os})~\cite{python3} for input/output and file system management. To keep the process modular, we implement separate webscraping scripts for each source (Kremlin vs.\ MID), parameterized by language. Each script follows the same two-stage pipeline shown in Figure~\ref{fig:webscrape-pipeline}. First, an index builder traverses the site’s public listings, records the canonical speech URLs and their numeric identifiers, and writes an index CSV. We rely on this index rather than naive “next page’’ crawling because listings can reorder, contain gaps, or include unpublished identifiers, and numeric IDs are not guaranteed to be contiguous. The index CSV therefore defines a stable target universe, enables resumable runs, and allows us to verify coverage. 

Second, a page fetcher-parser consumes the index CSV, downloads each speech page, applies conservative selectors to recover title, date, full text, some other metadata, and in the same pass discovers image sources exposed on the page or linked first-party photo subpages. For each candidate image we request the largest available rendition (falling back to smaller variants when necessary) and record both the number of images advertised on the page and the number successfully saved for that speech in the CSV so that any mismatch is explicit and auditable. Figure~\ref{fig:webscrape-pipeline} summarizes this two-stage workflow from site listings through to speech-level CSVs and image folders.

All crawlers are single-threaded and use fixed headers, bounded retries, and randomized inter-request delays to reduce load on the origin servers and to remain within expected norms of polite scraping. Because the index already enumerates the target universe, the page stage can safely skip any speech that is already present in the corresponding site-language CSV. This minimizes redundant requests and allows clean resumption after interruptions. For both sources (Kremlin and MID) we treat the numeric identifier embedded in each speech URL as the primary key within that site. This identifier is shared across that site’s Russian and English mirrors, which allows us to link Russian and English versions of the same speech by a simple equality join on \texttt{ID} rather than using fuzzy matching on speech titles or dates. Images are stored in per-speech folders named by this identifier, and files within each folder follow a simple sequence-based naming convention. Given a single CSV row, the corresponding image paths are therefore direct and unambiguous.

\begin{figure}[H]
    \centering
    \begin{tikzpicture}[
        font=\small,
        >=Latex,
        node distance=2.0cm and 3.0cm,
        table/.style={
            rectangle,draw,rounded corners,
            fill=gray!5,
            align=center,
            minimum width=4.6cm,
            minimum height=1.2cm
        },
        storage/.style={
            rectangle,draw,
            fill=gray!5,
            align=center,
            minimum width=6.0cm,
            minimum height=1.1cm
        },
        note/.style={
            rectangle,draw,dashed,
            align=left,
            minimum width=5.2cm,
            minimum height=1.2cm
        },
        arrow/.style={->,thick}
    ]

    \node[table] (en) {Source--EN CSV\\(e.g.\ Kremlin--EN)\\\texttt{ID}, date, title\_en, full\_text\_en, \dots};
    \node[table, right=of en] (ru) {Source--RU CSV\\(e.g.\ Kremlin--RU)\\\texttt{ID}, date, title\_ru, full\_text\_ru, \dots};

    \node[storage, below=2.0cm of $(en)!0.5!(ru)$] (imgroot)
        {Images root directory\\\texttt{IMAGES\_ROOT/\textit{ID}/} with sequence-numbered files};

    \node[note, below=0.8cm of imgroot] (example)
        {\small Example: \texttt{ID = 1185}\\
         \small EN and RU rows join on \texttt{ID}\\
         \small Images stored in \texttt{IMAGES\_ROOT/1185/}};

    \draw[arrow] (en.south) |- node[left,pos=0.45]{\texttt{ID}} (imgroot.north west);
    \draw[arrow] (ru.south) |- node[right,pos=0.45]{\texttt{ID}} (imgroot.north east);

    \end{tikzpicture}
    \caption{Cross-lingual linkage within a source. For each source (Kremlin or MID), Russian and English speech tables share the same numeric \texttt{ID} column, which serves as the primary key and is also used as the per-speech image folder name. Aligning Russian and English speeches, and linking them to their images, is therefore a simple equality join on \texttt{ID}, with no need for fuzzy matching on titles or dates.}
    \label{fig:crosslingual-linkage}
\end{figure}
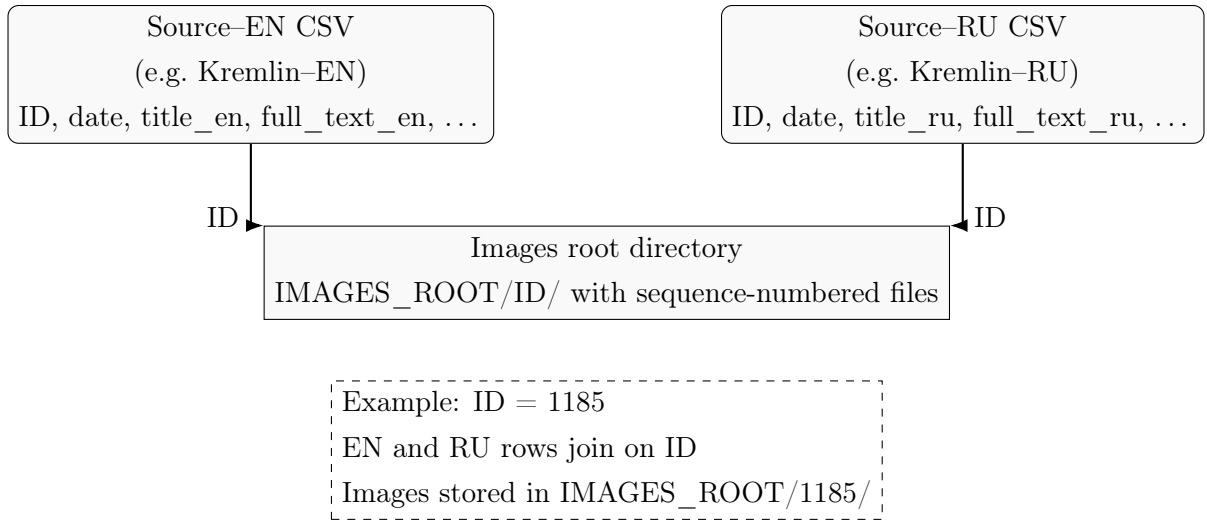

We keep the text body as close to the source as possible. The parser targets the main content region for each site and (Russian or English) language but otherwise avoids heavy boilerplate removal or aggressive normalization, preserving original punctuation, orthography, and capitalization in both Russian and English. Where source pages contain empty bodies or unusually long speeches, we retain these edge cases in the CSV rather than filtering or truncating them, so that corpus coverage remains transparent. For each site-language pair, the final outputs consist of a single speech-level CSV and an accompanying images root in which one subdirectory per speech identifier holds the associated image files. Within each source (Kremlin or MID), Russian and English speech tables can be aligned exactly on \texttt{ID}, and the same identifier is used as the folder name for the corresponding per–speech image directory. Figure~\ref{fig:crosslingual-linkage} illustrates how the Russian- and English-language datasets (texts and images) are linked via this shared identifier scheme.

The above discussion outlines the shared, reproducible pipeline from indexing to page and image capture to unified outputs across all four corpora. That being said, there are several unique steps associated with our Kremlin and MID webpages and corresponding speech and image content. The following subsections describe these source–specific details for the Kremlin and MID collections in turn.

\subsubsection*{Webscraping the Kremlin Corpus}

Complementing the shared pipeline outlined above, we use several Kremlin corpus-specific steps to construct parallel corpora of Russian- and English-language presidential speeches for the Kremlin by webscraping the official Kremlin transcript archives.

For Russian-language content, we scraped the Kremlin transcript archive at
\nolinkurl{kremlin.ru/events/president/transcripts/}; for English-language content, we scraped
the parallel archive at \nolinkurl{en.kremlin.ru/events/president/transcripts/}. In both cases,
data collection followed the same two-stage pipeline: (i) constructing a stable index of transcript
identifiers and URLs, and (ii) harvesting speech-level metadata, full text, and associated images
for each indexed item.

\paragraph{Index construction.}
For each language, we first crawl the Kremlin transcript listings and write an index CSV of the form
\texttt{page number, id, url}. Each index entry records the listing page number, a numeric transcript identifier, and the canonical absolute URL for the transcript. The indexer uses a realistic browser user-agent and \texttt{Accept-Language} headers, enforces request timeouts, and implements conservative back-off logic for HTTP 403/429/5xx error responses. We retain only entries whose URLs match the expected transcript pattern, so that every downstream row in the final corpus can be keyed by a stable numeric identifier via the \texttt{id} column.

\paragraph{HTTP session management and politeness.}
The transcript harvester reads the specific Kremlin (English and Russian) index file and iterates over the set of unique identifiers per language. All HTTP requests are routed through a session wrapper that centralizes headers, timeouts, and pacing. For each transcript we issue a lightweight \texttt{HEAD} request against the main transcript URL; if the server returns a 404 error, the identifier is marked as unavailable and skipped entirely. For surviving IDs, \texttt{GET} requests are spaced by 9--10 seconds between transcript pages and photo gallery pages, with exponential backoff and (for the Russian corpus) optional proxy rotation when encountering temporary blocks. We also check response headers and body size to ensure that only well-formed HTML pages are processed.

\paragraph{Speech-level metadata and text extraction.}
For each successfully retrieved transcript page, we parse a consistent set of speech-level metadata and map these fields directly into the final Kremlin CSV schema.

\setlength{\tabcolsep}{10pt}
\renewcommand{\arraystretch}{1.15}

\newcolumntype{L}[1]{>{\raggedright\arraybackslash}p{#1}}

Table~\ref{tab:kremlin_schema} documents the column names used in our processed Kremlin (English and Russian) metadata CSV files and their meanings. List-valued fields (e.g., declared tags, speakers, image filepaths, image-topic IDs) are stored as serialized list strings (e.g., \nolinkurl{["...","..."]}) and are blank when the information is unavailable.

\textbf{Language note:} The \emph{Kremlin English} metadata CSV does not contain all translation columns; fields ending in \nolinkurl{_english} appear only in the \emph{Kremlin Russian} metadata CSV and store English translations of the corresponding Russian-language variables.

\textbf{Probability storage note:} In our processed files, \nolinkurl{curated_topic_probability} is stored as a single scalar (top-1 confidence for the assigned text topic), while \nolinkurl{curated_image_topic_probabilities} is a list with one scalar confidence value per image.

\textbf{Caption availability note:} When images are present, image-topic lists are aligned with \nolinkurl{stored_image_filepaths}; image captions are extracted when available and may be missing for some images, resulting in occasional length mismatches between caption lists and the stored image list.

\begin{longtable}{@{}L{0.34\linewidth}@{\hspace{1.2em}}L{0.62\linewidth}@{}}
\caption{Kremlin speech-level columns in the processed metadata CSV files.}
\label{tab:kremlin_schema}\\

\textbf{Column} & \textbf{Description} \\
\hline
\endfirsthead

\textbf{Column} & \textbf{Description} \\
\hline
\endhead

\nolinkurl{id} & Numeric transcript identifier associated with the speech page. \\
\nolinkurl{url} & Final resolved URL of the transcript page. \\

\nolinkurl{title} & Speech title/headline as displayed on the transcript page. \\

\nolinkurl{title_english} & English translation of \nolinkurl{title} (Kremlin Russian CSV only). \\

\nolinkurl{full_text} & Full extracted speech body text. \\

\nolinkurl{full_text_english} & English translation of \nolinkurl{full_text}  (Kremlin Russian CSV only). \\

\nolinkurl{full_text_word_count} & Word count of the transcript text (computed as the number of whitespace-separated tokens from the extracted speech body). \\

\nolinkurl{date} & Human-readable date string as displayed on the webpage. \\
\nolinkurl{year} & Calendar year extracted from the parsed date/time metadata when available. \\
\nolinkurl{month} & Month name (stored as a full English month name, e.g., \emph{January}). \\
\nolinkurl{day} & Day of month (numeric). \\
\nolinkurl{time} & Clock time parsed from time metadata (blank if not provided). \\

\nolinkurl{location} & Location string as provided on the transcript page (raw/original form). \\
\nolinkurl{location_english} & English translation of \nolinkurl{location} (Kremlin Russian CSV only). \\
\nolinkurl{latitude} & Latitude in decimal degrees obtained by geocoding \nolinkurl{location} (blank if unresolved). \\
\nolinkurl{longitude} & Longitude in decimal degrees obtained by geocoding \nolinkurl{location} (blank if unresolved). \\

\nolinkurl{page_summary} & Short summary/lead text (from meta description or page intro blocks when available). \\
\nolinkurl{page_summary_english} & English translation of \nolinkurl{page_summary} (Kremlin Russian metadata CSV only). \\

\nolinkurl{speakers} & Extracted speaker name(s) associated with the transcript (serialized list string). \\

\nolinkurl{declared_geography} & Geography-related tags declared on the page (serialized list string). \\
\nolinkurl{declared_geography_english} & English translation of \nolinkurl{declared_geography} (Kremlin Russian metadata CSV only). \\

\nolinkurl{declared_topics} & Topic tags declared on the page (serialized list string). \\
\nolinkurl{declared_topics_english} & English translation of \nolinkurl{declared_topics} (Kremlin Russian metadata CSV only). \\

\nolinkurl{declared_persons} & Person/entity tags declared on the page (serialized list string). \\
\nolinkurl{declared_persons_english} & English translation of \nolinkurl{declared_persons} (Kremlin Russian metadata CSV only). \\

\nolinkurl{curated_topic_id} & Final (curated) text-topic identifier assigned to the transcript. \\
\nolinkurl{curated_text_topic_label} & Human-readable label for \nolinkurl{curated_topic_id}. \\
\nolinkurl{curated_text_topic_group} & Higher-level group/category for the curated text topic. \\
\nolinkurl{curated_topic_probability} & Top-1 text-topic probability (single scalar confidence value for the assigned \nolinkurl{curated_topic_id}). \\

\nolinkurl{stored_image_filepaths} & Local filepaths of downloaded images linked to the transcript (serialized list string). \\
\nolinkurl{saved_images_count} & Number of images successfully saved locally for the transcript. \\
\nolinkurl{declared_images_count} & Number of images declared on the transcript web page. \\
\nolinkurl{missing_images_count} & Number of missing images (\nolinkurl{declared_images_count} minus \nolinkurl{saved_images_count}). \\

\nolinkurl{image_captions} & Captions extracted for the transcript images in the original language when available (serialized list string; may be missing for some images). \\
\nolinkurl{image_captions_english} & English translation of \nolinkurl{image_captions} (Kremlin Russian metadata CSV only). \\

\nolinkurl{curated_image_topic_ids} & Assigned image-topic IDs (serialized list; aligned with \nolinkurl{stored_image_filepaths} when images are present). \\
\nolinkurl{curated_image_topic_labels} & Human-readable labels for the image topics (serialized list). \\
\nolinkurl{curated_image_group_names} & Higher-level group names for the image topics (serialized list). \\
\nolinkurl{curated_image_topic_probabilities} & Per-image top-1 probabilities (serialized list of floats; one probability score per image, aligned with \nolinkurl{stored_image_filepaths}). \\

\hline
\end{longtable}

\paragraph{Image discovery, de-duplication, and completeness.}

A core goal of the Kremlin corpus data collection stage is to pair each speech with the set of photographs displayed alongside it on the official site. To achieve this we use a combination of HTML pattern matching, content-based de-duplication, and cross-checking against the site's own photo counters.

For each transcript we inspect both the main transcript page and, when present, the corresponding photo gallery at the derived \texttt{/photos} URL. On each page we search for images using several strategies: (i) dedicated Kremlin slideshow containers, (ii) standard \texttt{<img>} tags (including lazy-loading attributes and \texttt{srcset} variants), (iii) images embedded in \texttt{<picture>} and \texttt{<noscript>} blocks, (iv) direct links whose \texttt{href} targets an image file, (v) inline background images declared in style attributes, and (vi) hero images exposed via Twitter/OpenGraph meta tags. For each candidate, we normalize its URL relative to the page base and construct a small set of plausible size variants (favoring high-resolution ``big2x'' or ``big'' versions, then smaller renditions, and finally the original URL as a fallback). URLs that clearly point to thumbnails (e.g., those containing \texttt{/thumb} or \texttt{/preview}) are discarded unless they are the only available version of the image.

We then download the binary content for each surviving candidate and compute a SHA-256~\cite{nist_fips180_4}. Within a transcript-level image directory we treat matching hashes as duplicates and retain only a single copy. This prevents multiple size variants or repeated gallery images from contaminating the corpus. Successful downloads are saved under language-specific image roots (e.g., \texttt{kremlin\_russian\_images/}) in subdirectories named by \texttt{id}. Within each subdirectory, filenames follow a consistent pattern of the form \texttt{<id>\_<seq>.ext}, where \texttt{<id>} is a Numeric transcript identifier associated with the speech page and \texttt{seq} is a sequential counter. The final CSV column \texttt{stored\_image\_filepaths} stores a JSON-encoded list of these local paths, while \texttt{image\_captions} stores the corresponding captions extracted from surrounding \texttt{<figcaption>} elements, caption blocks, or, when necessary, image \texttt{alt}/\texttt{title} attributes.

The Kremlin interface displays an explicit photo count for each speech (e.g., in a photo tab labeled with a number). Whenever such a counter is present, we parse it into the \texttt{declared\_photos} column. For all Kremlin English and Russian speeches in our final corpus,
\texttt{declared\_photos} is non-empty, and \texttt{missing\_photos\_count}
is identically zero, indicating that our harvested image counts exactly match the Kremlin interface wherever the site exposes a photo counter.

\paragraph{Final coverage and descriptive statistics.}
After scraping, cleaning, and merging, the final Kremlin English file contains \textbf{10{,}553} rows (one per speech), and the final Kremlin Russian file contains \textbf{13{,}340} rows. Both archives cover the same temporal span, from 31~December~1999 through 20~September~2025. All rows in both files have parseable dates: in the Russian corpus, 13{,}340/13{,}340 rows have valid \texttt{date} values, and in the English corpus, 10{,}553/10{,}553 rows have valid \texttt{date} values as well.

Text coverage is nearly complete. In the Russian file only six rows have an empty \texttt{full\_text} (and hence \texttt{word\_count} = 0), corresponding to purely image-based notices; the remaining 13{,}334 rows contain non-empty text. In the English file only four rows have empty \texttt{full\_text}. Across non-empty texts, Russian speeches have a maximum \texttt{word\_count} of 33{,}352, a mean of 1{,}904.67 words, and a median of 763. English speeches reach a maximum of 39{,}898 words, with a mean of 1{,}453.19 and a median of 806 words. In both languages, therefore, the corpus contains a mix of short statements and very long, policy-heavy addresses.

Image coverage is also substantial. In the Russian file, the total number of saved images is 44{,}779, with a minimum of 0, a maximum of 104, a mean of 3.36, and a median of 2 images per speech. A total of 3{,}605 Russian speeches have no associated images (\texttt{images\_count} = 0), while 9{,}735 speeches have at least one local image. In the English file, \texttt{images\_count} is non-null for all 10{,}553 rows; the total number of saved images is 38{,}637, with a minimum of 0, a maximum of 104, a mean of 3.66, and a median of 2 images per speech. Here 3{,}438 speeches lack images and 7{,}115 speeches have one or more images. In both languages, \texttt{declared\_photos} is non-empty for every row and \texttt{missing\_photos\_count} is identically zero, confirming that the final image counts track the Kremlin’s declared photo numbers exactly.

Kremlin transcript pages typically include an explicit event location field, which we extract and store in raw/original form as \texttt{location}. In the processed Russian Kremlin file, 12{,}684 of 13{,}340 rows have a non-empty \texttt{location} string (656 rows are empty). In the processed English Kremlin file, 9{,}830 of 10{,}553 rows have a non-empty \texttt{location} string (723 rows are empty). For downstream use, we also derive latitude and longitude coordinates from available location strings using the same geolocation transformation pipeline applied across both the Kremlin and MID corpora; full details, validation, and edge cases are provided in Section~\ref{subsec:location-geocoding} (Location Extraction and Geocoding Section further below).

Because the Kremlin reuses the same numeric identifier for parallel Russian- and English-language versions of a speech, we can align the two files exactly on \texttt{id}. The Russian file contains 13{,}340 unique identifiers and the English file contains 10{,}553; 10{,}094 identifiers appear in both files. This leaves 3{,}246 Russian-only speeches and 459 English-only speeches. Put differently, 75.7\% of Russian IDs have a matching English row, while 95.7\% of English IDs have a matching Russian row. 

\begin{figure}[H]
    \centering

    \begin{subfigure}[t]{0.48\textwidth}
        \centering
        \includegraphics[width=\linewidth]{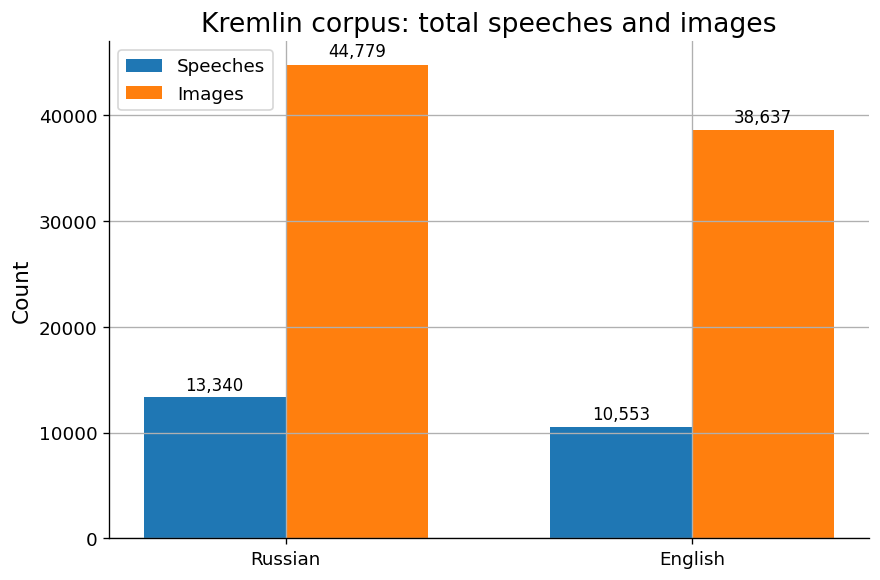}
        \caption{Total number of Kremlin speeches and harvested local images in the Russian- and English-language corpora.}
        \label{fig:kremlin_totals}
    \end{subfigure}
    \hfill
    \begin{subfigure}[t]{0.48\textwidth}
        \centering
        \includegraphics[width=\linewidth]{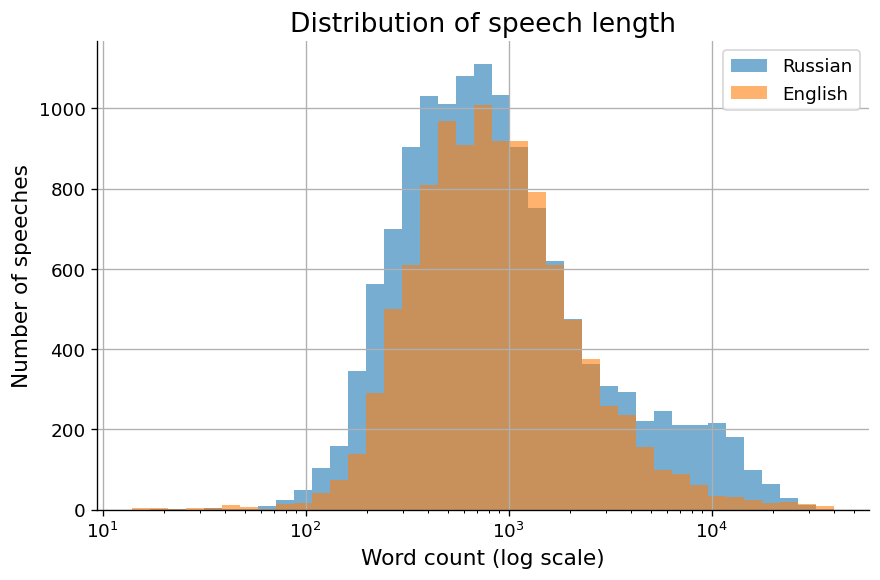}
        \caption{Distribution of Kremlin speech length (word count) for Russian- and English-language transcripts (log-scaled $x$-axis).}
        \label{fig:kremlin_wordcount}
    \end{subfigure}

    \vspace{0.8em}

    \begin{subfigure}[t]{0.48\textwidth}
        \centering
        \includegraphics[width=\linewidth]{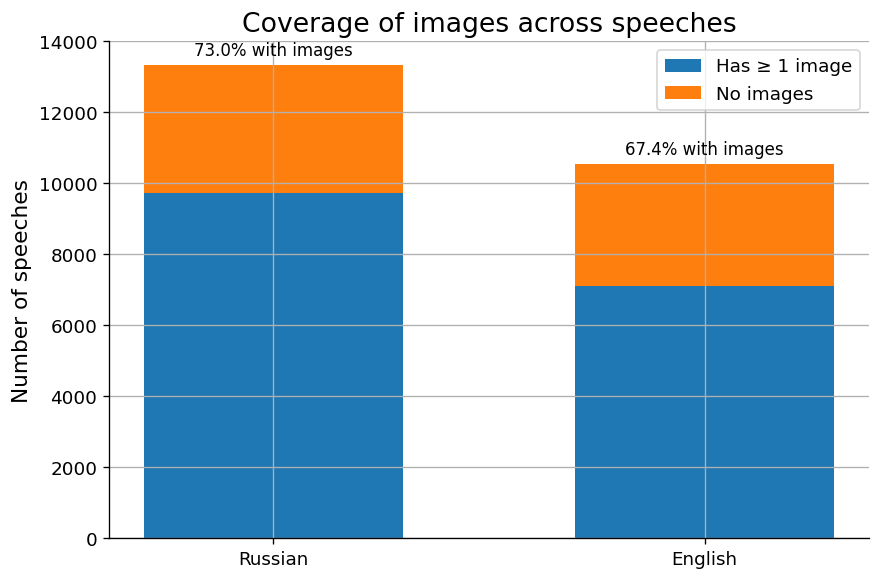}
        \caption{Share of Kremlin speeches with at least one local image versus no images, separately for the Russian- and English-language corpora.}
        \label{fig:kremlin_images_zero_vs_nonzero}
    \end{subfigure}
    \hfill
    \begin{subfigure}[t]{0.48\textwidth}
        \centering
        \includegraphics[width=\linewidth]{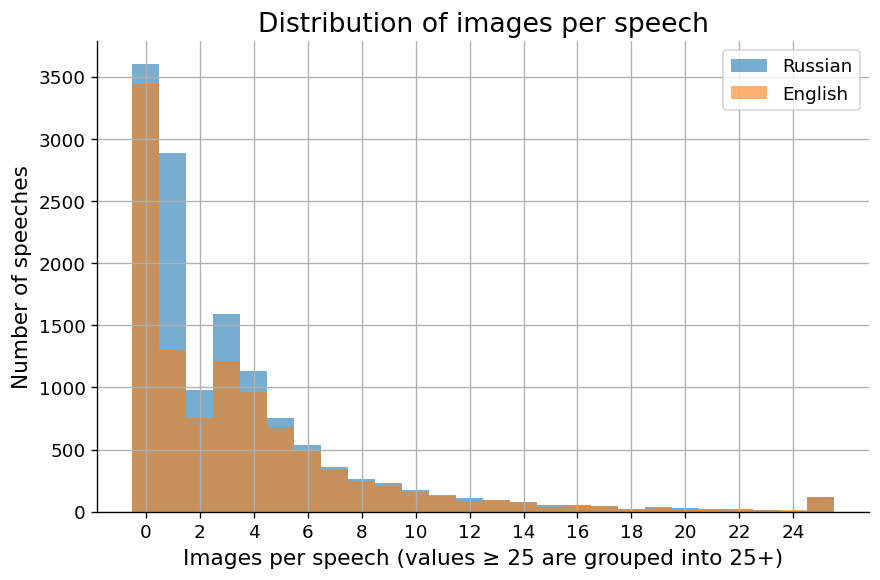}
        \caption{Distribution of images per Kremlin speech in the Russian- and English-language corpora (image counts $\geq 25$ are grouped into a single 25+ bin).}
        \label{fig:kremlin_images_per_speech}
    \end{subfigure}

    \caption{Kremlin corpus: coverage and basic descriptive statistics for speeches and images across Russian- and English-language versions of the site.}
    \label{fig:kremlin_overview}
\end{figure}

Figures~\ref{fig:kremlin_totals}--\ref{fig:kremlin_images_per_speech} summarize the coverage and basic structure of our scraped Kremlin speech corpus. Figure~\ref{fig:kremlin_totals} reports the total number of speeches and harvested local images in each language. The Russian-language archive contains 13{,}340 speeches and 44{,}779 images, while the English-language archive contains 10{,}553 speeches and 38{,}637 images. These totals highlight both the overall size of the corpus and the fact that the Russian site is somewhat more complete than its English counterpart, especially for earlier years.

Figure~\ref{fig:kremlin_wordcount} plots the distribution of \texttt{word\_count} for Russian- and English-language transcripts. We use a logarithmic scale on the $x$-axis because speech lengths are highly skewed and span more than two orders of magnitude: transcripts range from very short announcements to nearly 40{,}000-word events. On a linear scale, the dense middle of the distribution would be compressed and the long right tail would be almost invisible. The two language distributions overlap heavily, with most speeches falling between roughly $10^2$ and $10^4$ words, but the Russian distribution exhibits a heavier right tail. This indicates a larger number of very long Russian transcripts, consistent with the Russian site hosting additional full-length meetings, press conferences, and Q\&A sessions that are only partially translated, or not translated at all, on the English site.

Figure~\ref{fig:kremlin_images_zero_vs_nonzero} focuses on image coverage. Approximately 73\% of Russian-language speeches and 67\% of English-language speeches contain at least one locally stored image, with the remaining speeches consisting of text only. This confirms that images are pervasive but not universal in the Kremlin’s online communications. Importantly, for all pages where the source site reports an explicit image count, our stored image counts match exactly, indicating complete recovery of images for these image-bearing pages in both languages.

Finally, Figure~\ref{fig:kremlin_images_per_speech} reports the full distribution of \texttt{images\_count} per speech (for legibility, extremely image-rich pages with $\geq 25$ images are pooled into a single 25+ bin). Most speeches in both languages contain between zero and four images, with a long but thin tail of speeches that include ten or more images (for example, extended photo galleries associated with major public events). The similarity of the Russian and English distributions suggests that, conditional on a speech having at least one image, the Kremlin tends to mirror the basic image layout across language versions. Detailed variable descriptions and additional descriptive statistics are provided in the Data Records Section.

\subsubsection*{Webscraping the MID.RU Corpus}

We also tailor and refine our aforementioned shared pipeline for webscraping in order to construct parallel corpora of Russian- and English-language MID speeches, as obtained from the official MID website. For Russian MID speeches, we target the main ministerial speeches archive at \texttt{mid.ru/ru/press\_service/minister\_speeches/}, and for English we target its English-language counterpart at \texttt{mid.ru/en/press\_service/minister\_speeches/}. As with the Kremlin corpus, data collection proceeds in two stages: (i) building a stable index of transcript identifiers and URLs for each language, and (ii) harvesting speech-level metadata, text, images, and locations for every indexed transcript.

\paragraph{Index construction.}
For each language, we crawl the paginated listing of ministerial speeches by iterating over the \texttt{?PAGEN\_1=<page>} parameter. On each listing page we identify links to individual speeches using language-specific CSS selectors and a strict URL pattern that extracts the numeric identifier from paths of the form \texttt{/press\_service/minister\_speeches/<id>/} (or the corresponding English-language path with \texttt{/en/}). For each match we record the listing page number, the numeric ID, and the canonical absolute URL in an index CSV with columns \texttt{page\_num}, \texttt{id}, and \texttt{url}. IDs are de-duplicated across pages, and the crawler stops after two consecutive listing pages contain no new IDs. This ensures that we do not request empty or outdated archive pages.

\paragraph{HTTP session management and politeness.}
The MID harvester reads these index files and iterates over the set of unique IDs in each language (i.e., English and Russian). All requests are issued through a shared \texttt{requests}~\cite{requests2025} object that centralizes headers, timeouts, and pacing. To avoid overloading the MID servers, we enforce a conservative delay of five seconds between \emph{all} HTTP requests, including both transcript pages and image downloads, and apply standard timeouts and error handling. Responses are verified for status codes and basic content sanity before further processing. As with the Kremlin scraper, we treat any speech page that lacks a meaningful combination of title, text, or image(s) as an error and halt processing, so that every row in the final processed MID corpora corresponds to a page with substantive content.

\paragraph{Speech-level metadata and text extraction.}
For each successfully retrieved MID page, we parse a consistent set of speech-level metadata and map these fields directly into the final MID CSV schema. As in the Kremlin data, list-valued fields (e.g., stored image paths, captions, topic IDs, and probability vectors) are stored as serialized list strings (e.g., \nolinkurl{["...","..."]}) and are left blank when the corresponding information is unavailable for a given document.

\textbf{Language note:} The \emph{MID English} CSV and \emph{MID Russian} CSV share the same core schema. The only translation-specific column in the MID files is \nolinkurl{image_captions_english}, which appears \emph{only} in the \emph{MID Russian} CSV and stores English translations of the corresponding Russian-language \nolinkurl{image_captions} field.

\textbf{Probability storage note:} In our processed files, \nolinkurl{curated_topic_probability} is stored as a single scalar (top-1 confidence for the assigned text topic), while \nolinkurl{curated_image_topic_probabilities} is a list with one scalar confidence value per image.

\textbf{Caption availability note:} When images are present, image-topic lists are aligned with \nolinkurl{stored_image_filepaths}; image captions are extracted when available and may be missing for some images, resulting in occasional length mismatches between caption lists and the stored image list.

\setlength{\tabcolsep}{10pt}      
\renewcommand{\arraystretch}{1.15}
\newcolumntype{L}[1]{>{\raggedright\arraybackslash}p{#1}}

\begin{longtable}{p{0.28\linewidth}p{0.68\linewidth}}
\caption{MID data dictionary (speech-level CSVs).}
\label{tab:mid-data-dictionary}\\

\textbf{Column} & \textbf{Description} \\
\hline
\endfirsthead

\textbf{Column} & \textbf{Description} \\
\hline
\endhead

\nolinkurl{id} & Numeric document identifier associated with the MID page. \\
\nolinkurl{url} & Final resolved URL of the MID document page. \\

\nolinkurl{full_text} & Full extracted document body text (English in MID EN; original Cyrillic in MID RU$\rightarrow$EN). \\
\nolinkurl{full_text_english} & English translation of \nolinkurl{full_text} (MID RU$\rightarrow$EN CSV only; aligned row-for-row with the original Russian text). \\

\nolinkurl{full_text_word_count} & Word count of the document body text (computed as the number of whitespace-separated tokens from the extracted document body). \\

\nolinkurl{date} & Human-readable publication date string as displayed on the webpage. \\
\nolinkurl{year} & Calendar year extracted from the parsed date/time metadata when available. \\
\nolinkurl{month} & Month name (stored as a full English month name, e.g., \emph{January}). \\
\nolinkurl{day} & Day of month (numeric). \\
\nolinkurl{time} & Clock time parsed from page metadata (blank if not provided). \\

\nolinkurl{location} & Final location string for the document (blank if no clear event location can be recovered). \\
\nolinkurl{latitude} & Latitude in decimal degrees obtained by geocoding \nolinkurl{location} (blank/\texttt{NaN} if unresolved). \\
\nolinkurl{longitude} & Longitude in decimal degrees obtained by geocoding \nolinkurl{location} (blank/\texttt{NaN} if unresolved). \\

\nolinkurl{speakers} & Extracted speaker name(s) associated with the document (serialized list string). \\

\nolinkurl{curated_topic_id} & Final (curated) text-topic identifier assigned to the document from the MID topic space ($K=32$; integer in $\{0,\dots,31\}$). \\
\nolinkurl{curated_text_topic_label} & Human-readable label for \nolinkurl{curated_topic_id}. \\
\nolinkurl{curated_text_topic_group} & Higher-level group/category for the curated text topic. \\
\nolinkurl{curated_topic_probability} & Top-1 text-topic probability (single scalar confidence value for the assigned \nolinkurl{curated_topic_id}). \\

\nolinkurl{stored_image_filepaths} & Local filepaths of downloaded images linked to the document (serialized list string; stored as relative paths with respect to the MID corpus image root directory in the data release). \\
\nolinkurl{saved_images_count} & Number of images successfully saved locally for the document. \\
\nolinkurl{declared_images_count} & Number of images declared on the MID webpage when available (blank if not provided). \\
\nolinkurl{missing_images_count} & Number of missing images (\nolinkurl{declared_images_count} minus \nolinkurl{saved_images_count}). \\

\nolinkurl{image_captions} & Captions extracted for the document images in the original language when available (serialized list string; one caption per image, aligned with \nolinkurl{stored_image_filepaths}). \\
\nolinkurl{image_captions_english} & English translation of \nolinkurl{image_captions} (MID RU$\rightarrow$EN CSV only; serialized list aligned with \nolinkurl{stored_image_filepaths}). \\

\nolinkurl{curated_image_topic_ids} & Assigned image-topic IDs (serialized list; aligned with \nolinkurl{stored_image_filepaths} when images are present). \\
\nolinkurl{curated_image_topic_labels} & Human-readable labels for the image topics (serialized list; aligned with images). \\
\nolinkurl{curated_image_group_names} & Higher-level group names for the image topics (serialized list; aligned with images). \\
\nolinkurl{curated_image_topic_probabilities} & Per-image top-1 probabilities (serialized list of floats; one probability score per image, aligned with \nolinkurl{stored_image_filepaths}). \\

\hline
\end{longtable}

\paragraph{Image discovery and photo counters.}
The design of the MID's ministerial speeches section differs from the Kremlin site in that images associated with a MID speech are concentrated in a dedicated photo-album widget. For each MID speech we therefore restrict image harvesting to the ``Photo album'' or ``Additional materials'' area, identified by the \texttt{\#photo-slider .photo-slider\_\_list} container. Within this slider we enumerate all \texttt{<li>} elements, extract the associated \texttt{<img>} tags, and resolve their image URLs relative to the page base. For each candidate we record the best available source URL (using \texttt{src}, \texttt{data-src}, or the first entry in \texttt{srcset}) and the corresponding caption, taken from the image’s \texttt{alt} attribute where available.

Images are downloaded to language-specific directories (e.g., \texttt{mid\_russian\_scraped\_images/} and \texttt{mid\_english\_scraped\_images/}) in subfolders named by \texttt{id}. Within each subfolder, filenames follow a simple pattern \texttt{<id>\_<seq>.<ext>}, where \texttt{seq} is a zero-based counter and \texttt{ext} is the original file extension when present (otherwise \texttt{.jpg}). The final processed MID CSVs store these paths as JSON-encoded lists in the \texttt{stored\_image\_filepaths} column and the associated captions in \texttt{image\_captions}. To avoid trivial duplicates we de-duplicate images within each speech by the basename of their source URL.

As on the Kremlin site, \texttt{mid.ru} exposes a per-speech photo counter in the photo-album interface. We parse this counter into \texttt{declared\_photos}, falling back to the number of images in the slider list when the counter text is not present. The realized number of downloaded images for each speech is recorded in \texttt{images\_count}, and we store the difference between the declared and realized counts in \texttt{missing\_photos\_count}. In the final MID Russian and English processed files, \texttt{declared\_photos} is non-empty for every row and \texttt{missing\_photos\_count} is identically zero, indicating that for all speeches the number of locally stored images exactly matches the \texttt{mid.ru} photo counter.

\paragraph{Location recovery and geocoding.}
Because MID.RU does not consistently report venues in a dedicated structured field, we recover a usable \texttt{location} string using a conservative multi-stage pipeline that preserves observed locations when present and backfills only when missing. In brief, we (i) extract locations directly from MID page headers/titles when explicitly stated, (ii) apply lightweight rule-based parsing and named-entity recognition on the English representation of the record (title and full text) for remaining blanks, and (iii) run a final backfill pass using \texttt{Anthropic Claude 3 Haiku} (\texttt{claude-3-haiku-20240307}) to return a single location phrase or \texttt{UNKNOWN} when no venue is stated. We then geocode unique non-empty locations using a two-stage cascade (Nominatim first, ArcGIS fallback) implemented in \texttt{geopy}, storing \texttt{latitude} and \texttt{longitude} when resolution succeeds and leaving coordinates missing otherwise. Full implementation details, edge cases, and additional validation checks are reported in Section~\ref{subsec:location-geocoding} (Location Extraction and Geocoding Section further below).

\paragraph{Final coverage and descriptive statistics.}
After scraping, cleaning, and merging, the final Russian MID corpus contains \textbf{6{,}056} speeches and the final English MID corpus contains \textbf{5{,}057} speeches. Together these files span the period from 18~March~2004 through early October~2025: Russian-language speeches run from 18~March~2004 to 9~October~2025, while English-language speeches extend from 18~March~2004 to 7~October~2025. All rows in both files have parseable dates.

Text coverage in MID is high but not quite complete. In the Russian MID file only seven rows have an empty \texttt{full\_text} (and hence \texttt{word\_count} = 0); the remaining 6{,}049 rows contain non-empty text. In the English MID file only one row has empty \texttt{full\_text}. Among speeches with non-empty text, Russian MID transcripts have a maximum \texttt{word\_count} of 16{,}834, a mean of 1{,}099.41 words, and a median of 612.5, while English MID transcripts reach a maximum of 19{,}895 words, with a mean of 1{,}392.26 and a median of 786 words. Thus, compared to the Kremlin corpora, MID speeches are somewhat shorter on average, and the English MID texts tend to be slightly longer than their Russian counterparts, potentially reflecting both translation effects and differences in how the Ministry edits its Russian and English releases.

Image coverage is substantial but somewhat lighter than for the Kremlin corpora. In the Russian MID file, \texttt{images\_count} is non-null for all 6{,}056 rows, with a total of 4{,}498 saved images. The minimum number of images per speech is 0 and the maximum is 19; the mean number of images per speech is 0.74 and the median is 1.0. A total of 2{,}150 Russian MID speeches (35.5\%) have no associated images, while 3{,}906 speeches (64.5\%) have at least one local image. In the English MID file, \texttt{images\_count} is again non-null for all 5{,}057 rows, with 4{,}145 total images; here the minimum is 0, the maximum is 20, the mean is 0.82, and the median is 1.0 image per speech. Among English speeches, 1{,}466 (29.0\%) lack images and 3{,}591 (71.0\%) include at least one local image. As noted above, \texttt{declared\_photos} is non-empty and \texttt{missing\_photos\_count} is zero for every MID speech, indicating that our harvested image counts exactly match the MID interface wherever \texttt{mid.ru} exposes a photo counter.

\begin{figure}[H]
    \centering

    \begin{subfigure}[t]{0.48\textwidth}
        \centering
        \includegraphics[width=\linewidth]{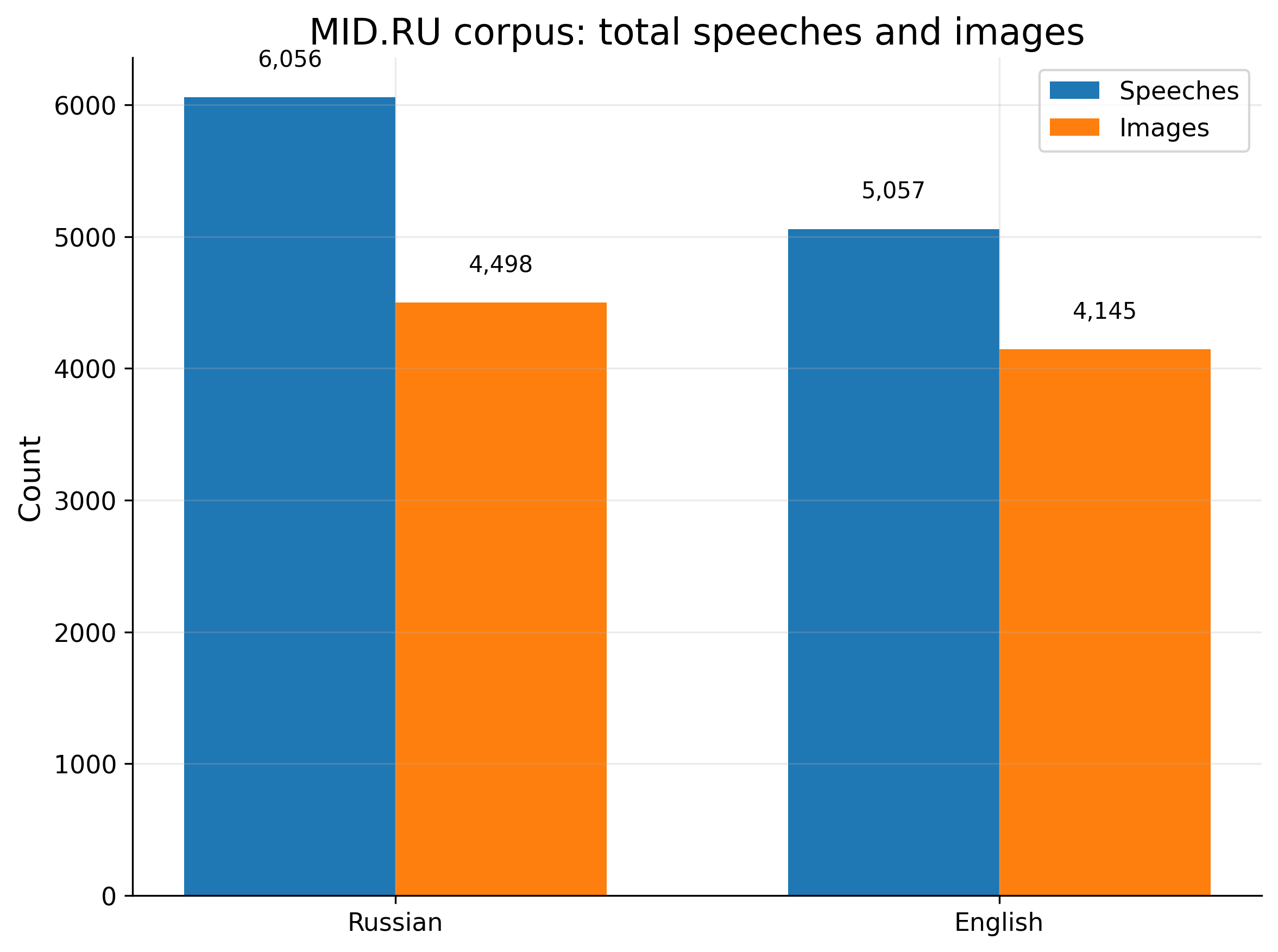}
        \caption{Total number of MID.RU speeches and harvested local images in the Russian- and English-language corpora.}
        \label{fig:mid_totals}
    \end{subfigure}
    \hfill
    \begin{subfigure}[t]{0.48\textwidth}
        \centering
        \includegraphics[width=\linewidth]{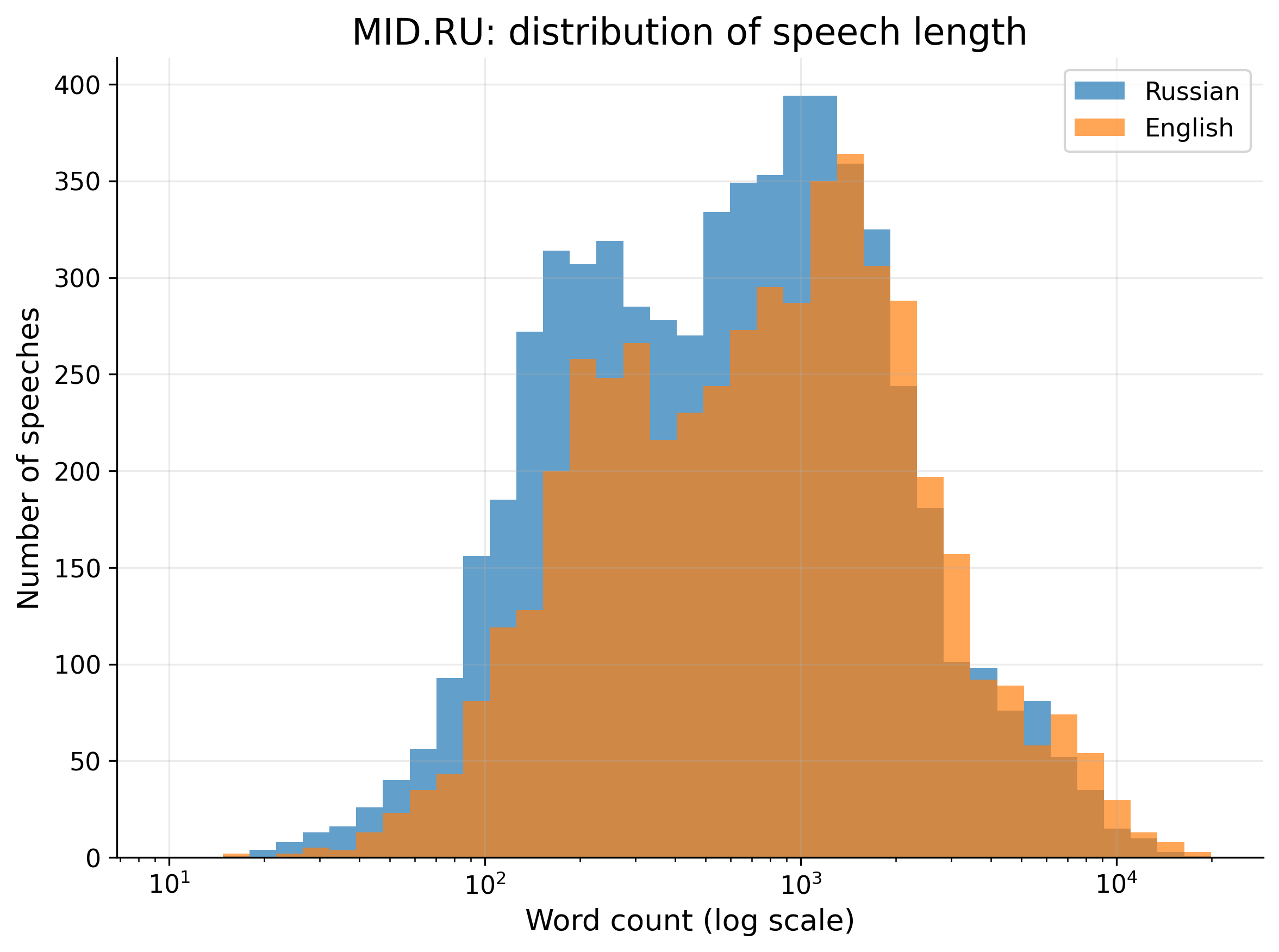}
        \caption{Distribution of MID.RU speech length (\texttt{word\_count}) for Russian- and English-language transcripts (log-scaled $x$-axis).}
        \label{fig:mid_wordcount}
    \end{subfigure}

    \vspace{0.8em}

    \begin{subfigure}[t]{0.48\textwidth}
        \centering
        \includegraphics[width=\linewidth]{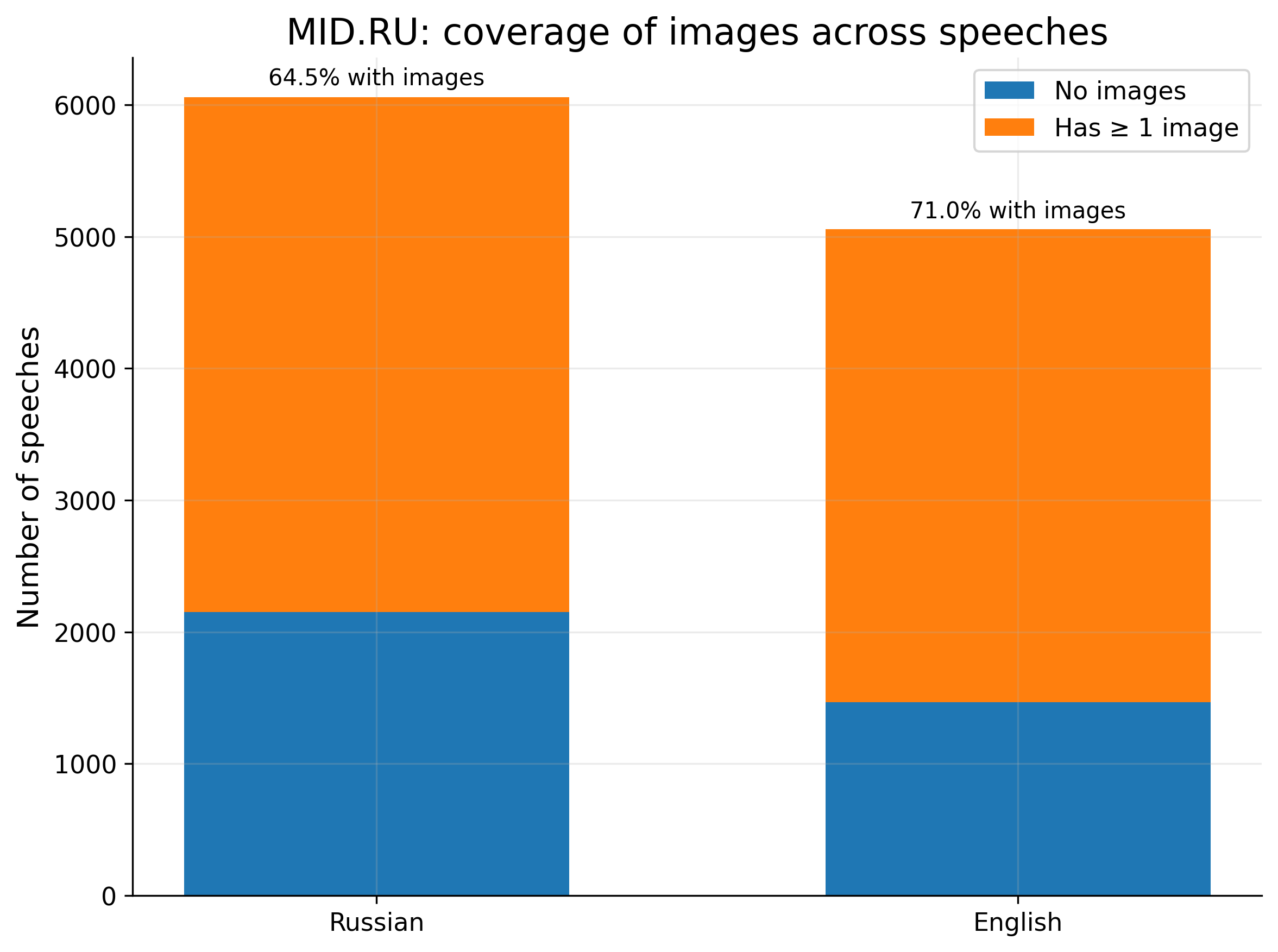}
        \caption{Share of MID.RU speeches with at least one local image versus no images, separately for the Russian- and English-language corpora.}
        \label{fig:mid_images_zero_vs_nonzero}
    \end{subfigure}
    \hfill
    \begin{subfigure}[t]{0.48\textwidth}
        \centering
        \includegraphics[width=\linewidth]{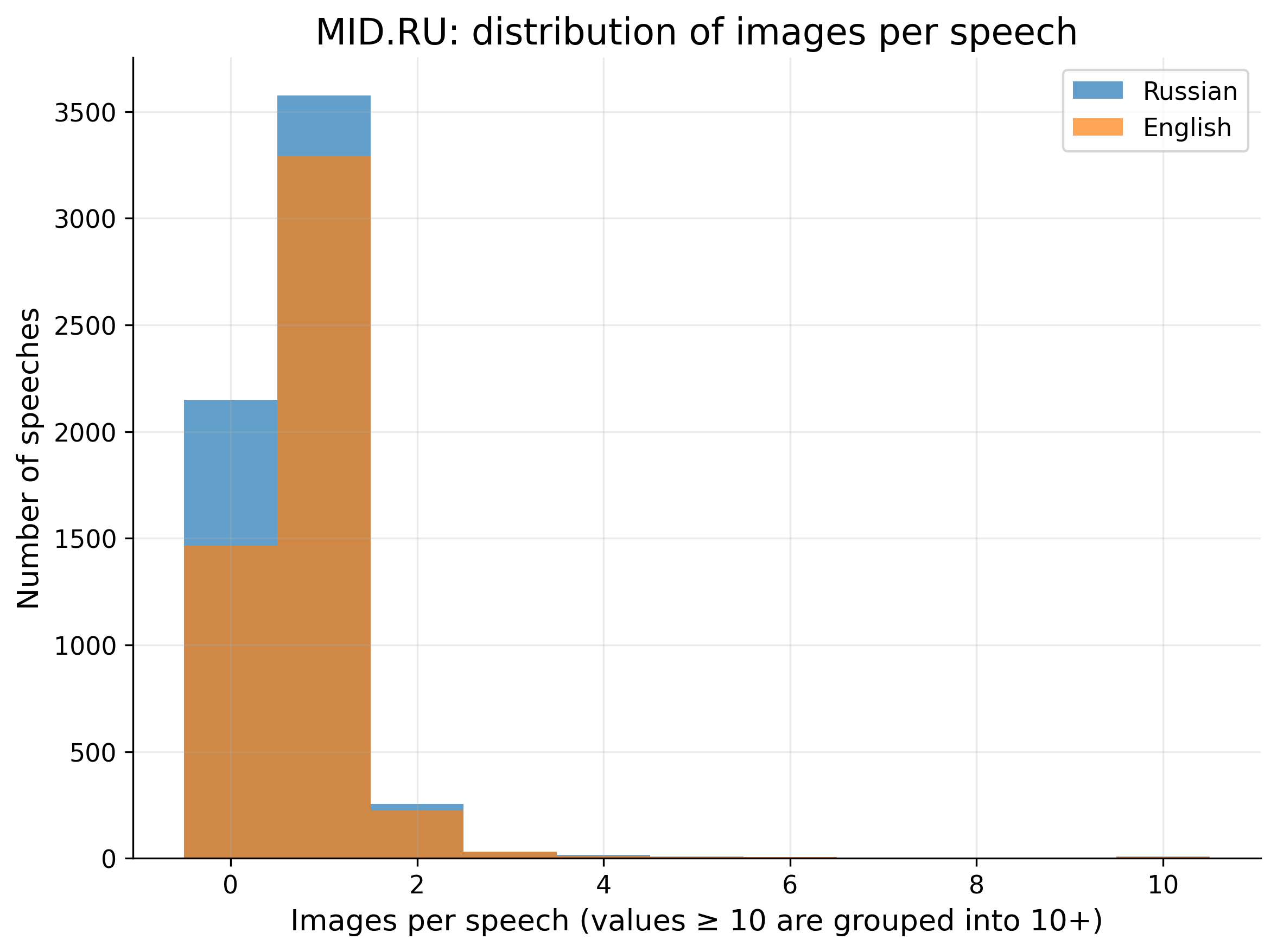}
        \caption{Distribution of images per MID.RU speech in the Russian- and English-language corpora (image counts $\geq 10$ are grouped into a single 10+ bin).}
        \label{fig:mid_images_per_speech}
    \end{subfigure}

    \caption{MID.RU corpus: coverage and basic descriptive statistics for speeches and images across Russian- and English-language versions of the site.}
    \label{fig:mid_overview}
\end{figure}

Figures~\ref{fig:mid_totals}--\ref{fig:mid_images_per_speech} summarize these coverage patterns for the MID corpora. Figure~\ref{fig:mid_totals} reports the total number of speeches and harvested local images in each language. The Russian MID archive contains 6{,}056 speeches and 4{,}498 images, while the English archive contains 5{,}057 speeches and 4{,}145 images. These totals highlight both the smaller scale of the MID corpora relative to the Kremlin corpora and the fact that, for MID, the number of English-language images is quite close to the number of Russian-language images despite the smaller number of English speeches.

Figure~\ref{fig:mid_wordcount} plots the distribution of \texttt{word\_count} for Russian- and English-language MID transcripts on a logarithmic scale. As with the Kremlin histograms, the log scale makes the long right tail of very long speeches visible while preserving detail in the denser middle of the distribution. Most MID speeches fall between roughly $10^2$ and $10^4$ words. The English distribution is slightly shifted to the right relative to the Russian distribution, reflecting the higher mean and median word counts in English; nonetheless, the two language distributions overlap heavily across the main range of speech lengths.

Figure~\ref{fig:mid_images_zero_vs_nonzero} focuses on image coverage. Roughly 64.5\% of Russian MID speeches and 71.0\% of English MID speeches contain at least one local image, with the remainder consisting of text-only releases. This pattern suggests that within the \texttt{mid.ru} archives, images are common but not universal, and that the English pages are slightly more likely than the Russian pages to include an accompanying photo album.

Finally, Figure~\ref{fig:mid_images_per_speech} shows the full distribution of \texttt{images\_count} per speech, pooling all speeches with ten or more images into a single 10$+$ bin for legibility. In both languages most speeches contain either zero or one image, and only a small minority of speeches host more than three images. Compared to the Kremlin distributions, the MID image counts are more tightly concentrated near zero and one, with very few large photo galleries. Together, these figures illustrate that the MID corpora provide broad longitudinal coverage of ministerial speeches with moderate but systematic photographic documentation, and suggest that the basic structure of speech lengths and image counts is similar across Russian and English versions of the site.

\subsection*{Additional Variables}

\subsubsection{Additional Extracted or Transformed Variables}
\label{subsec:additional-variables}

Beyond the core scraped text and page metadata, our final speech-level CSVs also include additional variables derived from the raw HTML and downstream processing pipelines (translation, location recovery, geocoding, and topic modeling). These variables are constructed in a harmonized way across all four corpora, with certain English-parallel fields present only for the Russian$\rightarrow$English translation files, as noted below.

\vspace{0.5em}
\noindent\textbf{Text fields.}
The \texttt{title} column stores the page headline in its original language (English for the EN corpora; Russian for the RU corpora). For the Russian corpora we additionally provide an Argos Translate~\cite{argos_translate} English rendering in \texttt{title\_english}. The main body text is stored as \texttt{full\_text} in the original language for all corpora; for the Russian files we also provide \texttt{full\_text\_english}, generated via our chunked Argos Translate~\cite{argos_translate} pipeline (long Russian texts are split into $\leq$4{,}000-character segments, translated segment-by-segment with basic quality checks, and concatenated back into a single English string). These translated fields are strictly row-aligned with the original Russian: \texttt{full\_text} and \texttt{full\_text\_english} always refer to the same speech-level document.

\vspace{0.5em}
\noindent\textbf{Dates, times, and derived calendar variables.}
The original publication date and time are stored in \texttt{date} and \texttt{time}, parsed from the source site HTML. From \texttt{date} we derive integer calendar variables \texttt{year}, \texttt{month}, and \texttt{day} (uniformly defined across corpora as four-digit year, month 1--12, day 1--31). The \texttt{time} field preserves the reported clock time (when present) as a simple string. These derived variables support time-series and panel construction without requiring users to re-parse the original date formats.

\vspace{0.5em}
\noindent\textbf{Locations, English location labels, and geocoordinates.}
Each CSV includes an event-location string in \texttt{location}, stored as a cleaned human-readable value (light whitespace and punctuation normalization). For the Kremlin corpora, \texttt{location} is scraped directly from the source site (we do not infer missing locations). For the Kremlin Russian$\rightarrow$English corpus, we additionally provide an English-rendered location string in \texttt{location\_english} while retaining the Russian original in \texttt{location}. For the MID corpora, many records lack an explicit location on the source site; we therefore apply a conservative recovery procedure that fills \texttt{location} only when the title and/or body text provide clear evidence of a venue, otherwise leaving the field blank (such that no location is invented without textual support). Approximate coordinates are provided in \texttt{latitude} and \texttt{longitude}. Full details of MID location recovery, Kremlin handling, and the geocoding cascade are provided in Section~\ref{subsec:location-geocoding} (Location Extraction and Geocoding Section further below).

\vspace{0.5em}
\noindent\textbf{Speaker names.}
We provide a speech-level \texttt{speakers} field constructed via corpus-specific extraction procedures tailored to how the Kremlin and MID sites encode speaker cues (structured HTML labels versus transcript-like prefixes). We then normalize extracted labels into canonical person names for analysis. Full extraction, normalization, and verification details are provided in Section~\ref{subsec:speaker-names}.

\vspace{0.5em}
\noindent\textbf{Document length and media count variables.}
We include simple length and count variables derived from the scraped page content:
\begin{itemize}
    \item \texttt{full\_text\_word\_count}: an integer count of whitespace-delimited tokens in the cleaned \texttt{full\_text} field, providing a crude but useful measure of speech length for filtering and modeling.
    \item \texttt{saved\_images\_count}: the number of images successfully downloaded and stored for the speech.
    \item \texttt{declared\_images\_count}: for the Kremlin corpora, the number of photos the source page claims to contain (a site-provided count rather than a crawler-derived count).
    \item \texttt{missing\_images\_count}: defined as \texttt{declared\_images\_count} minus \linebreak\texttt{saved\_images\_count}. This quantity is zero when all site-declared images were successfully scraped and positive when some declared images could not be downloaded (e.g., broken links or access restrictions). It is left missing when no declared image count is available.
\end{itemize}

\vspace{0.5em}
\noindent\textbf{Stored image paths and captions.}
For every speech we provide list-valued columns summarizing its associated images:
\begin{itemize}
    \item \texttt{stored\_image\_filepaths}: a serialized list of relative filepaths for each image successfully downloaded for that speech, linking the speech-level CSV to the underlying image files used in downstream multimodal processing.
    \item \texttt{image\_captions}: a parallel list of captions aligned with \texttt{stored\_image\_filepaths}. Captions are scraped from the HTML (e.g., figure captions or alt text) and lightly cleaned; when no caption is available, the corresponding entry is left empty.
    \item \texttt{image\_captions\_english} (Russian$\rightarrow$English corpora only): an English-rendered caption list aligned to \texttt{image\_captions}, produced via the same translation approach used for other Russian$\rightarrow$English metadata fields.
\end{itemize}

\vspace{0.5em}
\noindent\textbf{Page summaries (Kremlin corpora).}
For the Kremlin corpora we include a short page-level summary field in \texttt{page\_summary}. For the Kremlin Russian$\rightarrow$English corpus we additionally provide an English-rendered version in \texttt{page\_summary\_english}. These summaries are included as descriptive metadata to support browsing and quick inspection; they are not required for or included in topic estimation.

\vspace{0.5em}
\noindent\textbf{Site-declared tags (Kremlin only).}
In the two Kremlin CSVs we retain structured content tags provided by the Kremlin and its official website. The primary tag field is \texttt{declared\_topics}, which records any site-assigned topical labels for various thematic categories. We also retain additional site-declared metadata fields, including \texttt{declared\_geography} and \texttt{declared\_persons}. For the Kremlin Russian$\rightarrow$English corpus we additionally provide English-rendered versions of these tag fields (\texttt{declared\_topics\_english}, \texttt{declared\_geography\_english}, and \linebreak\texttt{declared\_persons\_english}). These site-declared annotations are not used to fit our \texttt{BERTopic}~\cite{grootendorst2022bertopic} models, but they support comparisons between the source site’s tagging scheme and our unsupervised topics, and can be leveraged for validation or supervised extensions---including in our own validations further below.

\vspace{0.5em}
\noindent\textbf{Final text and image topic variables (curated).}
Finally, we attach the curated text and image topic variables from Section~\ref{subsec:speech-image-topics} directly to each row. For text, we store:
\begin{itemize}
    \item \texttt{curated\_topic\_id}: the integer ID of the primary text topic for the speech, defined as $k^\ast(d) = \arg\max_k p(k \mid d)$ after reassigning any \texttt{hdbscan}~\cite{campello2013hdbscan} outliers to their nearest topic centroid. In the Kremlin corpora this lies in $\{0,\dots,88\}$; in the MID corpora it lies in $\{0,\dots,31\}$.
    \item \texttt{curated\_text\_topic\_label}: the short human-readable label assigned to that topic.
    \item \texttt{curated\_text\_topic\_group}: a broader group label capturing higher-level domains.
    \item \texttt{curated\_topic\_probability}: a serialized vector containing the full document--topic probability distribution $\{p(k\mid d)\}_{k=0}^{K-1}$ aligned to the topic IDs for that corpus.
\end{itemize}

On the image side, each speech stores list-valued image-topic summaries aligned with \linebreak\texttt{stored\_image\_filepaths}:
\begin{itemize}
    \item \texttt{curated\_image\_topic\_ids}: for each stored image, the ID of the topic most strongly associated with that image.
    \item \texttt{curated\_image\_topic\_labels}: the corresponding short topic labels in the same order as \texttt{curated\_image\_topic\_ids}.
    \item \texttt{curated\_image\_group\_names}: the corresponding group labels (again aligned in order).
    \item \texttt{curated\_image\_topic\_probabilities}: for each stored image, a serialized vector containing its full image--topic score/probability distribution aligned to the topic IDs for that corpus.
\end{itemize}

Because image-topic fields are defined at the level of individual images but stored as list-valued columns at the speech level in our CSVs, they can be used as speech-level summaries (e.g., whether any attached image is associated with a topic) or as a starting point for constructing an image-level dataset by exploding the lists.

Taken together, these additional extracted and transformed variables---translated text and metadata fields, calendar variables, extracted speakers, cleaned locations and geocoordinates, length and media counts, stored image paths and captions, site-declared tags, and curated text and image topic variables (including full probability vectors)---make the final CSV files immediately usable as analysis-ready datasets.

\subsubsection{Speech and Image Topic Labels}

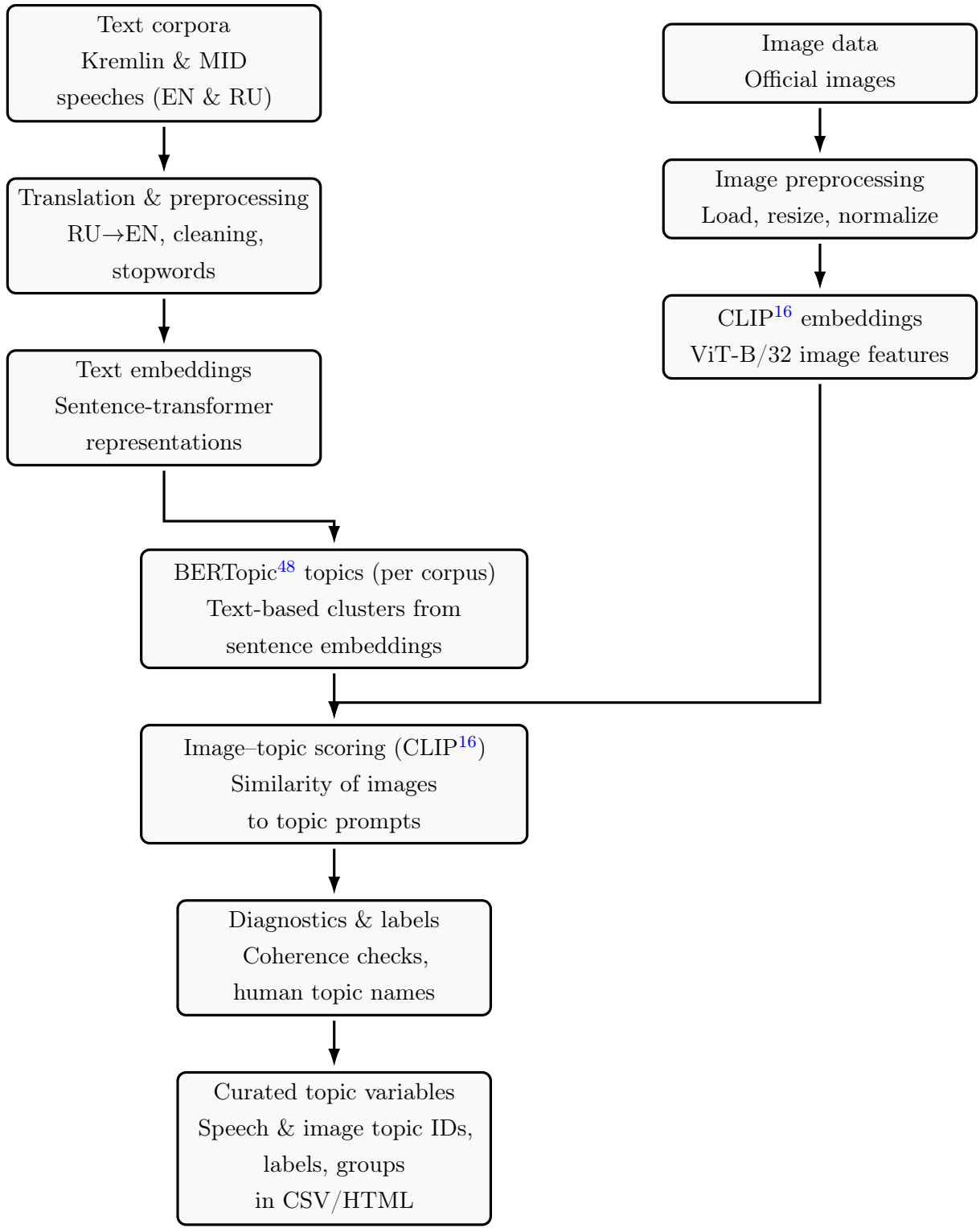
\begin{figure}[H]
\centering
\small 

\begin{tikzpicture}[
  node distance=0.9cm and 3.0cm,
  stage/.style={
    rectangle,
    rounded corners,
    draw=black,
    very thick,
    text width=4.8cm,
    minimum height=0.85cm,
    align=center,
    fill=gray!5,
    inner sep=5pt
  },
  line/.style={
    -{Latex[length=3mm,width=2mm]},
    very thick,
    shorten <=2pt,
    shorten >=2pt
  }
]

\node[stage] (t1) {\textbf{Text corpora}\\Kremlin \& MID\\speeches (EN \& RU)};
\node[stage, below=of t1] (t2) {\textbf{Translation \& preprocessing}\\RU$\rightarrow$EN, cleaning,\\stopwords};
\node[stage, below=of t2] (t3) {\textbf{Text embeddings}\\Sentence-transformer\\representations};

\node[stage, right=5.6cm of t1] (i1) {\textbf{Image data}\\Official images};
\node[stage, below=of i1] (i2) {\textbf{Image preprocessing}\\Load, resize, normalize};
\node[stage, below=of i2] (i3) {\textbf{CLIP~\cite{radford2021learning} embeddings}\\ViT-B/32 image features};

\node[stage, text width=6.0cm, below=1.35cm of t3, xshift=2.8cm] (j1)
{\textbf{\mbox{\texttt{BERTopic}~\cite{grootendorst2022bertopic} topics (per corpus)}}\\Text-based clusters from\\sentence embeddings};

\node[stage, text width=6.0cm, below=of j1] (j2)
{\textbf{\mbox{Image--topic scoring (CLIP~\cite{radford2021learning})}}\\Similarity of images\\to topic prompts};

\node[stage, below=of j2] (j3)
{\textbf{Diagnostics \& labels}\\Coherence checks,\\human topic names};

\node[stage, below=of j3] (j4)
{\textbf{Curated topic variables}\\Speech \& image topic IDs,\\labels, groups in CSV/HTML};

\draw[line] (t1) -- (t2);
\draw[line] (t2) -- (t3);

\draw[line] (i1) -- (i2);
\draw[line] (i2) -- (i3);

\coordinate (tDrop) at ($(j1.north)+(0,0.45cm)$);
\draw[line]
  (t3.south) -- (t3.south |- tDrop) -- (j1.north |- tDrop) -- (j1.north);

\draw[line] (j1) -- (j2);

\coordinate (belowJ1) at ($(j1.south)+(0,-0.55cm)$);
\draw[line]
  (i3.south) -- (i3.south |- belowJ1) -- (j2.north |- belowJ1) -- (j2.north);

\draw[line] (j2) -- (j3);
\draw[line] (j3) -- (j4);

\end{tikzpicture}

\caption{High-level overview of the topic modeling pipeline. We embed Kremlin and MID speech texts (EN and RU$\rightarrow$EN) with sentence-transformer models and fit \texttt{BERTopic}~\cite{grootendorst2022bertopic} separately per corpus. In parallel, associated images are embedded with CLIP~\cite{radford2021learning} (ViT-B/32~\cite{radford2021learning}) and scored against topic prompts to assign image-topic labels. Final curated topic IDs, labels, and groups are saved for speeches and images.}

\label{fig:bertopic-highlevel}
\end{figure}

This section discusses our topic modeling-derived variables, which assign each speech and its associated images interpretable topic labels. Figure~\ref{fig:bertopic-highlevel} provides a high-level overview of the topic-labeling pipeline used to produce these variables.

\label{subsec:speech-image-topics}

While our Kremlin corpora included a set of Kremlin-declared thematic (i.e., topic) tags, these were incomplete, accompanying only 62.2\% (6563/10553) and 62.2\% (8298/13340) of our English and Russian Kremlin speeches, respectively. We also do not know how these declared topics were assigned by the Kremlin itself, and our MID corpora do not include any MID-declared topic tags. In order to consistently assign thematic labels and hence topic variables to \textit{all} relevant speeches within our English and Russian Kremlin and MID corpora, we implement a text$+$image topic extraction pipeline followed by human labeling.

In this regard, our goal in this step is to turn each speech and its associated images into a set of interpretable, comparable topic labels that future researchers can use as covariates in downstream analyses. We achieve this by estimating topic models on four corpora that jointly cover the Russian presidency and the Ministry of Foreign Affairs (MID): Kremlin English speeches (Kremlin EN), Kremlin Russian speeches translated into English (Kremlin RU$\rightarrow$EN), MID English texts (MID EN), and MID Russian texts translated into English (MID RU$\rightarrow$EN). For each institution, we construct a single shared \emph{text-based} topic space (with $K=89$ topics for the Kremlin and $K=32$ for MID) that we apply across languages (native English and Russian$\rightarrow$English). We then use CLIP~\cite{radford2021learning}  (Contrastive Language--Image Pretraining; \cite{radford2021learning}) ---
a pretrained vision--language model that learns a shared embedding space for images and text---to map images into these same topic spaces, so that images can be associated with the learned topics without ever influencing the underlying text model. Concretely, CLIP~\cite{radford2021learning} produces a numeric vector (embedding) for each image and for each short text prompt; when an image and a prompt are semantically related, their embeddings are close (high cosine similarity). We therefore represent each topic with a short English prompt built from that topic's key terms (e.g., ``official photo about Ukrainian affairs''), encode both the prompts and images with CLIP~\cite{radford2021learning}, and assign images to topics by comparing similarities across all topic prompts. This procedure lets us score and label images using the same topic inventory learned from text, while keeping the topic model itself purely text-based. The resulting topic IDs, labels, and overarching topic group labels are stored directly in our final CSV files and serve as additional variables in our merged dataset. We next describe this approach in greater detail.

\vspace{0.5em}
\noindent\textbf{Software stack and computing environment.}
All text and image processing, embedding, and topic modeling is implemented in \texttt{Python 3}~\cite{python3} within\texttt{ Google Colab Pro+}~\cite{google_colab} notebooks. We rely on \texttt{pandas}~\cite{mckinney2010pandas} and \texttt{numpy}~\cite{harris2020numpy} for data handling, \texttt{PyTorch}~\cite{paszke2019pytorch} and \texttt{transformers}~\cite{wolf2020transformers} for neural models, \texttt{sentence-transformers}~\cite{reimers2019sentence} for sentence-level embedding models, \texttt{BERTopic}~\cite{grootendorst2022bertopic} for topic modeling, \texttt{umap-learn}~\cite{mcinnes2018umap} for dimensionality reduction, and \texttt{hdbscan}~\cite{campello2013hdbscan} for density-based clustering.

\noindent
Although the final topic models are estimated on English-language text (including Russian$\rightarrow$English translations for the Russian corpora), we also conducted extensive exploratory experiments that attempted to model topics directly from Cyrillic Russian text. These experiments did not yield sufficiently stable or interpretable solutions for inclusion in the final pipeline, but they informed our preprocessing choices for the Russian corpora. For these Russian-specific experiments and preprocessing steps, we use \texttt{Stanza}~\cite{qi2020stanza}, \texttt{spaCy}~\cite{spacy} (with a large Russian model), and \texttt{pymorphy3}~\cite{pymorphy3} for morphological normalization.

\noindent
Image handling uses \texttt{Pillow}~\cite{pillow}, and CLIP models~\cite{radford2021learning} are accessed via \texttt{sentence-transformers}~\cite{reimers2019sentence}. Embedding and topic-model estimation runs are executed on NVIDIA A100 GPUs provided by Google Colab Pro+~\cite{google_colab}, while heavier Russian preprocessing and the direct-Russian topic-modeling experiments are CPU-bound and executed on high-RAM Colab Pro+ instances (on the order of tens of CPU-hours per Russian corpus). We use \texttt{joblib}~\cite{joblib} for parallelization and on-disk \texttt{Parquet}~\cite{apache_parquet} shards to stream documents and avoid exhausting RAM.

\vspace{0.75em}
\noindent\textbf{Pipeline overview (released vs.\ diagnostic components).}
We describe first the \emph{released} topic-labeling workflow, which defines the canonical topic IDs and curated topic variables distributed in the CSVs. This released workflow operates in a shared \emph{English semantic space}: it uses native-English texts for the English corpora and machine-translated English text fields for the Russian corpora, enabling direct cross-language comparability within each institution. 
We also implemented an earlier \emph{native-Russian (Cyrillic) topic-modeling pipeline} that attempted to fit \texttt{BERTopic}~\cite{grootendorst2022bertopic} directly on the original Russian texts (rather than on translated English). This workflow required substantially heavier Russian-specific preprocessing (sentence segmentation, Cyrillic/noise gating, morphological normalization, and expanded Russian stopwords) and extensive experimentation with multiple multilingual embedding models and tuning choices. Despite these efforts, the resulting native-Russian topic solutions were not sufficiently coherent and stable for inclusion as the released topic inventories. We therefore adopt a translation-based strategy for the final datasets: we translate the Russian corpora into English (RU$\rightarrow$EN), apply the same English-side preprocessing, and fit \texttt{BERTopic}~\cite{grootendorst2022bertopic} in English space for the Kremlin RU$\rightarrow$EN and MID RU$\rightarrow$EN corpora. The released topic IDs and curated topic labels are thus derived from the English and translated-English pipelines, while the native-Russian pipeline is retained only as an exploratory workflow.

\vspace{0.75em}
\noindent\textbf{Released (English-space) workflow.}
For each institution (Kremlin and MID), the released workflow proceeds in five steps:
(i) text cleanup (and RU$\rightarrow$EN translation for Russian corpora),
(ii) English-space document embeddings,
(iii) \texttt{BERTopic}~\cite{grootendorst2022bertopic} estimation and stabilization of a shared topic inventory per institution,
(iv) image--topic scoring using CLIP~\cite{radford2021learning} against the fixed text-topic inventory,
and (v) human annotation of topic labels and group labels, followed by writing curated variables to the final CSVs.

\vspace{0.75em}
\noindent\textbf{English-space text construction: RU$\rightarrow$EN translation and preprocessing.}
For all four corpora, our goal is to construct a clean, comparable \emph{English-space} text input for downstream topic modeling. We therefore (i) translate the Russian corpora into English to create parallel English text fields, and then (ii) apply the same English-side preprocessing steps to \emph{all} English-space text---both native English and translated English.

\emph{Russian corpora (Kremlin RU$\rightarrow$EN, MID RU$\rightarrow$EN): translation.}
We construct parallel English fields (\texttt{title\_english}, \texttt{full\_text\_english}) by translating each row’s original Cyrillic text using \texttt{Argos Translate}~\cite{argos_translate}, applied consistently across both Russian corpora. Translation is performed strictly row-by-row: \texttt{full\_text} (Cyrillic) and \texttt{full\_text\_english} always refer to the same speech, and we do not reorder, merge, split, or otherwise alter document boundaries during translation. IDs and URLs are preserved.

\emph{English-space preprocessing (applied uniformly to native and translated English).}
After translation, we treat \texttt{full\_text} (native English) and \texttt{full\_text\_english} (translated English) as a single class of English-space inputs and apply identical preprocessing steps. We start from HTML-stripped text and apply light normalization: Unicode whitespace normalization (including replacement of non-breaking spaces), removal of residual scraping artifacts, and simple punctuation cleanup (e.g., stripping redundant line breaks introduced by HTML extraction). We do not apply stemming or lemmatization to English-space text. This is deliberate: \texttt{BERTopic}~\cite{grootendorst2022bertopic}'s class-based TF--IDF representation (c-TF--IDF) operates on surface forms, and preserving proper nouns improves interpretability for political actors, organizations, and place names. Across both native-English and translated-English texts, we remove stopwords using a union of \texttt{spaCy}~\cite{spacy}'s default English stopword list and a small hand-curated set of highly frequent domain terms (e.g., ``russia'', ``russian federation'', ``president'') and year tokens (e.g., ``2008'') that otherwise dominate c-TF--IDF but add little topical information. The exact stopword lists are provided in the replication code and configuration files for this project.

\vspace{0.75em}
\noindent\textbf{English-space text embeddings (canonical).}
For all \emph{released} topic models (including the translated Russian corpora), we embed each speech using \texttt{sentence-transformers}~\cite{reimers2019sentence} with \linebreak
\texttt{all-mpnet-base-v2}~\cite{song2020mpnet}. Because many speeches exceed transformer context limits, we encode each document using overlapping windows of up to 512 word-piece tokens (a sliding window with small overlap to reduce boundary artifacts), producing one embedding per window. We L2-normalize window embeddings and average them to obtain a single 768-dimensional L2-normalized document vector per speech. For shorter speeches, this reduces to a single forward pass. These English-space embeddings (native English + translated English) define the semantic space used for the released topic IDs.

\vspace{0.75em}
\noindent\textbf{\texttt{BERTopic} modeling and shared topic inventories.}
We use \texttt{BERTopic}~\cite{grootendorst2022bertopic} to obtain topic assignments and \texttt{c-TF--IDF} topic representations from the English-space embeddings. \texttt{BERTopic}~\cite{grootendorst2022bertopic} combines dimensionality reduction (we use \texttt{umap-learn}~\cite{mcinnes2018umap}) with \texttt{HDBSCAN clustering}~\cite{campello2013hdbscan} and \texttt{c-TF--IDF} keyword extraction. Images never influence clustering; the topic model is fit purely on text embeddings.

We estimate four corpus-level \texttt{BERTopic}~\cite{grootendorst2022bertopic} models (Kremlin EN, Kremlin RU$\rightarrow$EN, MID EN, MID RU$\rightarrow$EN). In contrast to a single pooled model per institution, we fit \emph{one model per corpus}, and we treat the resulting topic IDs as \emph{corpus-specific} (i.e., topic 12 in Kremlin EN is not assumed to be the same substantive topic as topic 12 in Kremlin RU$\rightarrow$EN). We nonetheless enforce institution-level comparability in \emph{granularity} by fixing the same target topic count within each institution (Kremlin: $K=89$; MID: $K=32$), so that the English and translated corpora within the same institution are summarized at a similar level of detail.

\begin{figure}[H]
    \centering
    \includegraphics[width=\linewidth,height=0.45\textheight,keepaspectratio]{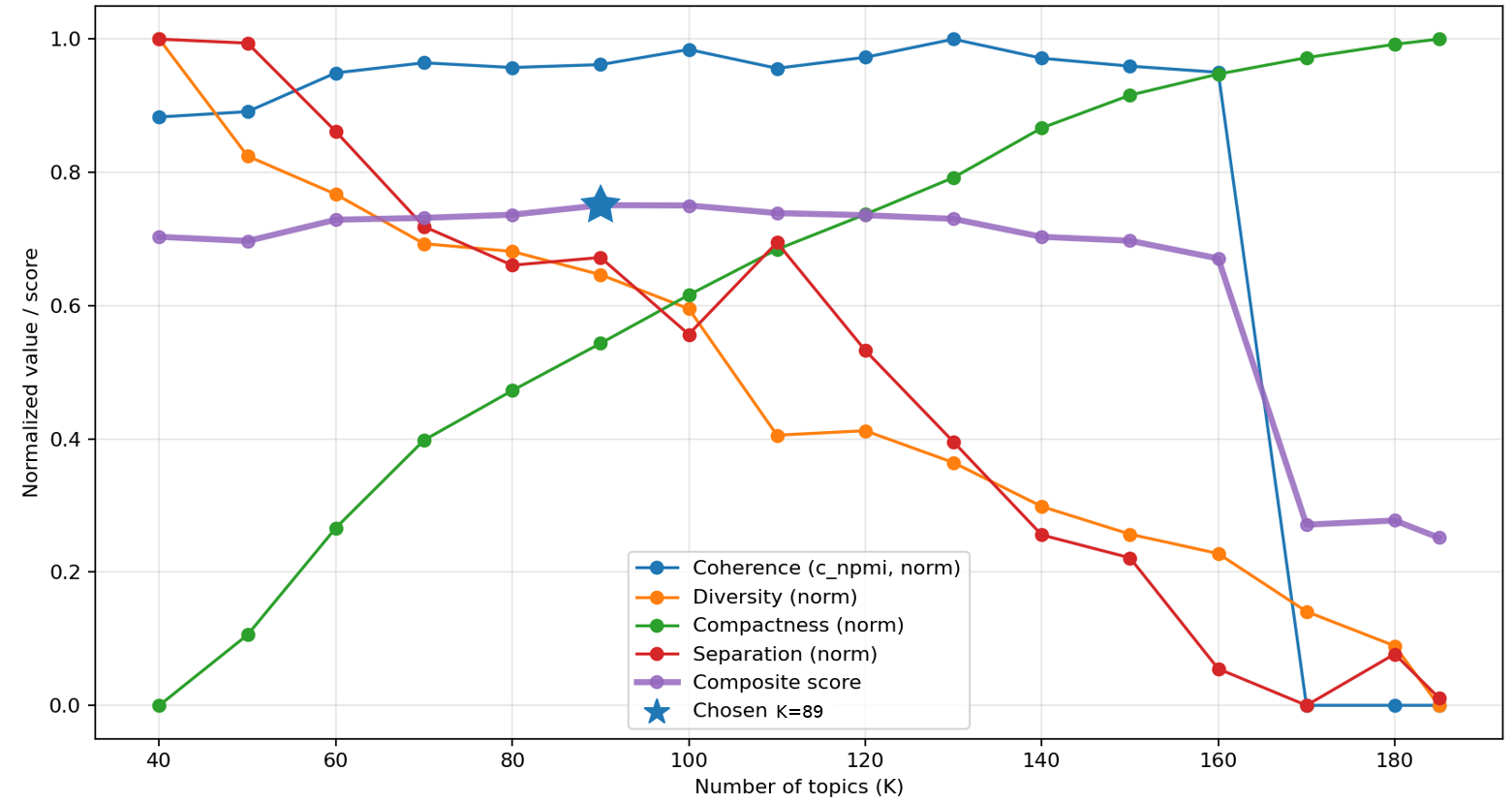}
    \caption{K-sweep scree plot for the Kremlin English corpus. Lines show normalized topic-quality metrics (coherence $c_{\mathrm{npmi}}$, diversity, compactness, and separation) and their weighted composite score; the selected solution is $K=89$ topics.}
    \label{fig:kremlin_k_sweep_90}
\end{figure}

\begin{figure}[H]
    \centering
    \includegraphics[width=\linewidth,height=0.45\textheight,keepaspectratio]{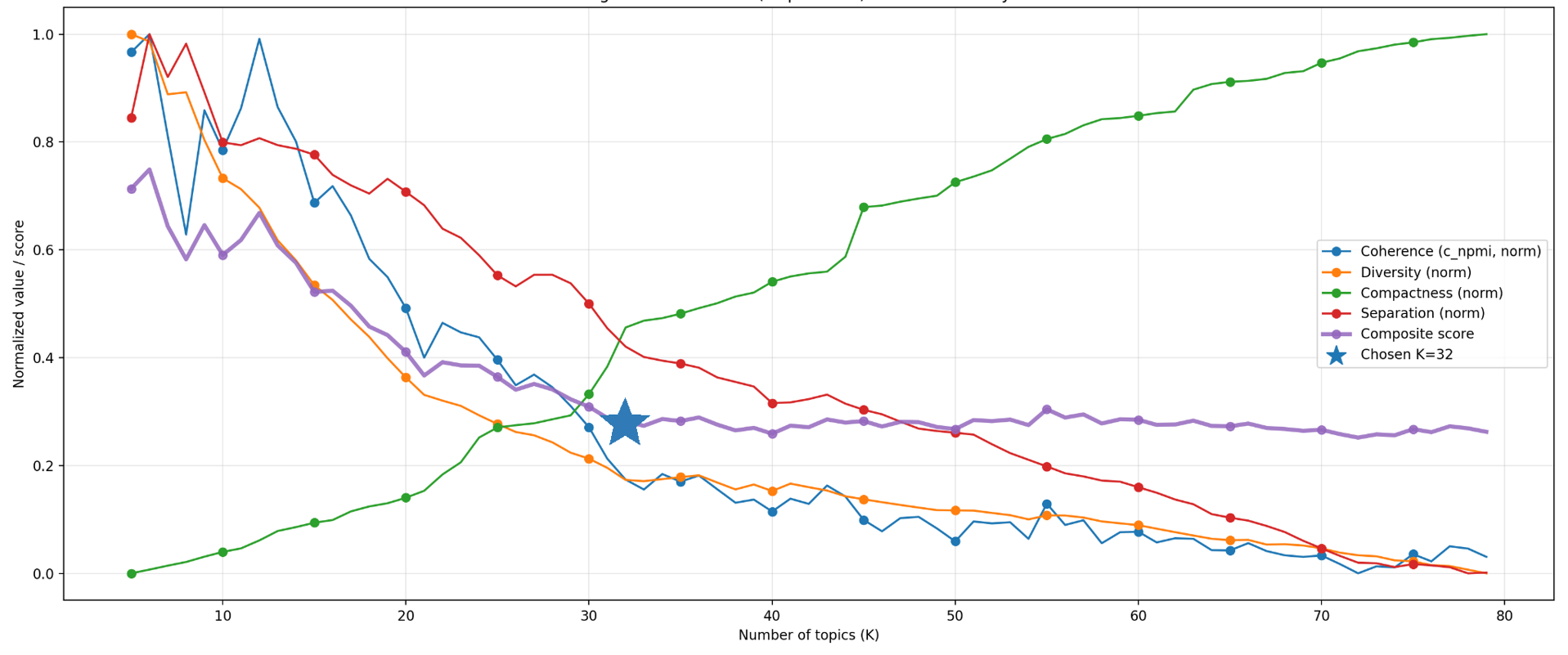}
    \caption{K-sweep scree plot for the MID English corpus. Lines show normalized topic-quality metrics (coherence $c_{\mathrm{npmi}}$, diversity, compactness, and separation) and their weighted composite score; the selected solution is $K=32$ topics.}
    \label{fig:mid_k_sweep_32}
\end{figure}

To select these target topic counts, we first performed model-selection diagnostics on the native-English corpora (Kremlin EN and MID EN). For each corpus, we began from a single high-resolution \texttt{BERTopic} fit (excluding \texttt{Topic} $=-1$ outliers throughout), which yielded up to 185 substantive topics for Kremlin EN and up to 79 substantive topics for MID EN. We then reduced each high-resolution solution to a grid of candidate topic counts and evaluated each candidate using four complementary criteria: semantic coherence (c\_NPMI), keyword diversity, within-topic compactness, and between-topic separation. In parallel, we monitored cluster diagnostics (topic-size distributions and outlier rates) and centroid-based redundancy checks (pairwise cosine similarities between topic centroids to detect near-duplicate topics). We summarize the four primary criteria with a weighted composite score (coherence 0.40, diversity 0.20, compactness 0.25, separation 0.15) and corroborate the quantitative diagnostics with qualitative inspection of each topic’s \emph{top 10} c-TF--IDF keywords and representative speeches to avoid overly coarse topics (too few $K$) or fragmented/redundant topic sets (too many $K$). This procedure favored $K=89$ for Kremlin EN and $K=32$ for MID EN as stable, interpretable solutions that balance thematic granularity against noise and redundancy. For consistency across each institution's language variants, we then fixed these same target topic counts when estimating the corresponding RU$\rightarrow$EN models (Kremlin RU$\rightarrow$EN and MID RU$\rightarrow$EN). As an additional face-validity check, we verified that the resulting keyword profiles in the Kremlin EN and Kremlin RU$\rightarrow$EN models were broadly similar at these target values, supporting the use of common topic granularity across languages even though topic IDs and human-assigned labels remain corpus-specific. We provide the full diagnostics outputs (the candidate-$K$ metric table and scree-style plots used for selection) in this paper's replication materials.

For each corpus, \texttt{BERTopic}~\cite{grootendorst2022bertopic} produces a ranked keyword representation for every topic (via c-TF--IDF) and a document--topic probability distribution. We then perform corpus-specific human annotation: for each topic in each corpus, we review its top 10 keywords and highest-probability speeches (and, where relevant, representative images from the associated HTML summaries) and assign (i) a short topic label and (ii) a broader topic-group label. These labels are therefore defined \emph{separately for each corpus} and are stored alongside the corpus-specific topic IDs in our final full CSVs.

\noindent\textbf{Outlier handling and primary topic assignment.}
\texttt{BERTopic}~\cite{grootendorst2022bertopic}/\texttt{HDBSCAN} may initially label some documents as outliers (topic $-1$). For production use we remove $-1$ labels by assigning each outlier to the nearest topic centroid (via cosine distance in embedding space). Drawing on applied topic model research in the social sciences \cite{bagozzi2015multifaceted,wilkerson2017large,berliner2021political}, we then define the primary (dominant) speech topic as the maximum a posteriori topic,
\[
k^\ast(d) = \arg\max_k p(k\mid d),
\]
and store this as \texttt{curated\_topic\_id}. Thus, in the final released CSVs, every speech has a valid \texttt{curated\_topic\_id} in $\{0,\dots,K-1\}$ with no remaining $-1$ labels. Once this text-topic inventory is fixed (including each topic’s c-TF--IDF \emph{top 10} keywords), we treat it as the canonical semantic reference for downstream image scoring and curation.

\vspace{0.75em}
\noindent\textbf{Image embeddings and image--topic scoring (CLIP).}
For each speech, we load all available images referenced in \texttt{stored\_image\_filepaths} using \texttt{Pillow}~\cite{pillow}, apply the standard CLIP preprocessing, and encode each image using a CLIP ViT-B/32 backbone~\cite{radford2021learning} (via \texttt{sentence-transformers}~\cite{reimers2019sentence}). This yields an L2-normalized image embedding for every stored image (images are treated individually; multi-image speeches retain multiple embeddings).

Crucially, CLIP does \emph{not} alter the fitted text topics. Instead, after the text-based topic inventory is fixed, we map images into that inventory using prompt-based similarity scoring. For each final topic $k$, we construct a short English prompt based on that topic’s top keywords (e.g., ``official photo about Ukrainian affairs'') and encode the prompt with the CLIP text encoder to obtain a topic vector $v^{\text{topic}}_k$. For each image $i$ with embedding $v^{\text{image}}_i$, we compute cosine similarities to all topic vectors and convert them into an image--topic probability distribution via a softmax:
\[
p(k \mid \text{image } i)
\;\propto\;
\exp\!\bigl(\lambda \cdot \cos(v^{\text{image}}_i, v^{\text{topic}}_k)\bigr),
\]
where $\lambda>0$ is a fixed temperature parameter (held constant across a corpus in the replication code). We assign each image a primary topic $k^\ast(i)=\arg\max_k p(k\mid i)$ and retain the full probability vectors for downstream robustness checks and qualitative inspection.

\vspace{0.75em}
\noindent\textbf{Human annotation of topic labels and groups, and handling assignment uncertainty.}
Once topic inventories and keyword lists are fixed, a Russian-speaking coauthor conducts an iterative qualitative coding pass over all topics. For each topic (within each corpus), the coder inspects the c-TF--IDF keyword list, the highest-probability speeches, and representative images (ranked using image--topic probabilities), and assigns (i) a short human-readable topic label (e.g., ``Ukrainian affairs'') and (ii) a broader group label capturing higher-level domains (e.g., ``Post-Soviet Relations'', ``Military \& Security'', ``Domestic Politics''). When upstream preprocessing refinements (e.g., improved stopword lists) change topic boundaries, labels are revisited to maintain internal consistency.

Because topics are learned by clustering in a continuous embedding space and because c-TF--IDF summarizes each topic using only its most distinctive surface keywords, labels should be interpreted as \emph{summary descriptors} rather than exhaustive tags. Substantively adjacent subthemes are often grouped together---especially in foreign-policy discourse---and a topic label may emphasize the most salient recurring referent even when the cluster contains closely related variants. For example, a topic whose top keywords include multiple East Asian referents (e.g., ``China'', ``Japan'', ``Beijing'', ``Tokyo'') may be labeled with the dominant descriptor (e.g., ``China'') even though some speeches within that cluster primarily concern Japan. This is an inherent trade-off when we distribute a tractable topic representation over many thousands of speeches: in the released speech-level tables we provide a single primary topic per speech for ease of use, while also providing the full item--topic probability information so users can model overlap and ambiguity when needed.

\vspace{0.75em}
\noindent\textbf{Writing curated variables and exporting full probability tables.}
For each speech, we store the dominant (highest-probability) topic as \texttt{curated\_topic\_id} and join this ID to the human-assigned topic label and group label, producing \texttt{curated\_text\_topic\_label} and \linebreak\texttt{curated\_text\_topic\_group}. For images, we store primary image-topic assignments and summarize them back into the speech-level CSV as list-valued columns (\texttt{curated\_image\_topic\_ids}, \texttt{curated\_image\_topic\_labels}, \texttt{curated\_image\_group\_names}), with one entry per stored image.

In addition to these dominant-topic variables, we export two auxiliary \emph{long-format} probability tables as supplemental files: (i) a document--topic table with one row per \texttt{speech\_id}$\times$\texttt{topic\_id} pair containing $p(k\mid d)$ for all $k\in\{0,\dots,K-1\}$, and (ii) an image--topic table with one row per \texttt{image\_id}$\times$\texttt{topic\_id} pair containing the corresponding image--topic score/probability for all topics. These long-format files allow users to (i) recover the full topic distribution for any speech or image, (ii) analyze mixtures rather than only the top topic (e.g., retaining the top-$m$ topics per speech), (iii) apply custom thresholds or uncertainty-aware models (e.g., filtering on $p(k\mid d)$), and (iv) conduct robustness checks beyond dominant-topic labeling.

\vspace{0.75em}
\noindent\textbf{Diagnostic native-Russian preprocessing and embedding pipeline.}
While not used in our final released topics, note that in parallel to the released English-space workflow, we invested substantial effort in a native-Russian pipeline operating directly on the original Cyrillic \texttt{full\_text}, with the goal of obtaining a fully Russian-language topic solution of comparable quality. In this case, we iterated through multiple increasingly stringent cleaning, normalization, and representation strategies and implemented a comparatively heavy preprocessing and embedding workflow designed specifically for Russian morphology and domain-specific boilerplate. Despite these efforts, and based upon a Russian-speaking expert's qualitative reviews, native-Russian topic solutions remained systematically less stable and less coherent than the English-space approach for our corpora. Accordingly, we do not release native-Russian clustering outputs as official topic IDs. We nonetheless document the pipeline here to demonstrate the extent of our diagnostic work and to provide a reproducible baseline for future improvements. In brief, the diagnostic pipeline:
(i) applies sentence segmentation with \texttt{Stanza}~\cite{qi2020stanza} and tokenization/POS/NER with \texttt{spaCy}~\cite{spacy},
(ii) implements ``Cyrillic gating'' to remove non-Russian noise (URLs, boilerplate, and lines dominated by Latin characters),
(iii) performs morphological normalization with \texttt{pymorphy3}~\cite{pymorphy3} (lemmatizing content-bearing tokens while preserving proper-noun surface forms where lemmatization harms interpretability),
(iv) applies an expanded Russian stopword list (including curated lists to remove frequent formalities and greeting formulas),
and (v) constructs a cleaned \texttt{model\_text} field per speech for downstream Russian embeddings and keyword extraction.

For Russian embeddings in this diagnostic workflow, we use an ensemble of multilingual embedding models (\texttt{intfloat/multilingual-e5-large}~\cite{wang2022e5}, LaBSE~\cite{feng2020labse}, and \texttt{BAAI/bge-m3}~\cite{bge_m3}) with the same 512-token windowing and normalization strategy used for English-space embeddings, and generated the top 10 keywords for each speech. In practice, however, these native-Russian experiments were not sufficiently stable or interpretable to be used in the released pipeline: topic keyword lists were often overly heterogeneous, topics did not align well with representative images, and small implementation/encoding artifacts could introduce mixed-language noise (e.g., English tokens appearing inside Cyrillic outputs). Qualitative feedback on these native-Russian outputs further emphasized that, without extensive per-topic document reading, many topics could not be reliably interpreted from keywords and images alone, suggesting that the model was not producing a coherent, analyst-usable topical structure for the MID RU corpus and that similar instability could appear in the Kremlin RU setting under alternative parameterizations. Because these issues undermine replicability and substantively meaningful topic labeling, we do \emph{not} use the native-Russian pipeline for any released topic IDs, curated topic variables, or keyword summaries.

We therefore used the earlier discussed translation-based strategy for the final datasets: we translate the Russian corpora into English (RU$\rightarrow$EN), apply the same English-side preprocessing used for the native-English corpora, and estimate the released \texttt{BERTopic}~\cite{grootendorst2022bertopic} models in a single English semantic space (native English + translated English). All released topic IDs and curated topic variables are derived exclusively from these English-space models.

\subsubsection{Location Extraction and Geocoding}
\label{subsec:location-geocoding}

We construct a harmonized set of location variables for all four corpora, consisting of a cleaned event location string (\texttt{location}) and approximate geographic coordinates (\texttt{latitude}, \texttt{longitude}). These fields support spatial aggregation, mapping, and distance-based analysis while preserving the original source information wherever possible.

\vspace{0.5em}
\noindent\textbf{Kremlin corpora (observed locations from the source site).}
For the Kremlin English corpus, event locations are scraped directly from the Kremlin website and stored in \texttt{location}. We do not infer or rewrite these strings beyond light whitespace and punctuation cleanup. For the Kremlin Russian$\rightarrow$English corpus, we retain the original Russian location text and additionally store an English-rendered version in \texttt{location\_english}. For geocoding and cross-corpus comparability, coordinates for the Kremlin Russian corpus are derived using \texttt{location\_english}.

\vspace{0.5em}
\noindent\textbf{MID corpora (location recovery using title \emph{and} full text, with an LLM backfill step).}
For both the MID English corpus and the MID Russian$\rightarrow$English translated corpus, \texttt{location} is the final named location column. We preserve any non-empty \texttt{location} values already present in the dataset and only attempt recovery when \texttt{location} is blank.

When \texttt{location} is blank, we recover a location using a \texttt{spaCy}~\cite{spacy}-based heuristic that uses both the document title and the full text. Specifically, we apply a fixed priority order:
(i) a rule-based extraction from the tail of the title (when the title ends with a location-like segment),
(ii) named-entity recognition (NER) on the title.
If no location is detected by any step, \texttt{location} remains blank.

After this \texttt{spaCy}~\cite{spacy} pass, we apply a second ``backfill'' pass for any remaining blanks using \texttt{Anthropic Claude 3 Haiku} (model identifier \texttt{claude-3-haiku-20240307})~\cite{anthropic_claude_haiku} via the Messages API. We prompt the model to extract \emph{only} the location phrase from the record’s English title and full text, and to return \texttt{UNKNOWN} when no location is clearly stated. We then write the returned location phrase into \texttt{location} for those rows (keeping blanks when the model returns \texttt{UNKNOWN}). Because the model may return either city-only strings (e.g., ``Moscow'') or composite strings (e.g., ``Moscow, Russia''), we standardize all recovered values after extraction by retaining only the city component (i.e., the substring before the first comma) and applying light whitespace/punctuation cleanup. As a result, the final MID \texttt{location} field is stored consistently in a city-only format.

The final MID CSVs do not retain intermediate extraction provenance fields (e.g., \texttt{location\_method}, \texttt{location\_confidence}, or raw NER candidate columns). Only the final \texttt{location} string and the geocoded coordinate columns (\texttt{latitude}, \texttt{longitude}) are included in the released dataset files.

\vspace{0.5em}
\noindent\textbf{Geocoding pipeline (all corpora).}
We map each unique non-empty \texttt{location} string to approximate coordinates using a two-stage geocoding cascade implemented with \texttt{geopy}~\cite{geopy}. We first query \texttt{Nominatim (OpenStreetMap)}~\cite{nominatim,openstreetmap}; if the request fails or times out, we fall back to the \texttt{ArcGIS geocoder}~\cite{esri_arcgis_geocoding}. We do not apply explicit ambiguity filtering or country-bias restrictions (e.g., Russia-only constraints) and accept the first geocoder result returned. Coordinates are stored as raw floating-point values (no rounding), and any location that cannot be resolved by either service remains missing (\texttt{NaN}) in \texttt{latitude}/\texttt{longitude}. We do not distribute a separate persistent geocode lookup table; only the final per-row coordinate columns are included in the replication package.

\vspace{0.5em}
\noindent\textbf{Coverage in the final datasets.}
Table~\ref{tab:location-coverage} summarizes location and coordinate coverage in the final corpora. ``Loc'' indicates non-empty \texttt{location}; ``Coords'' indicates both \texttt{latitude} and \texttt{longitude} are non-missing; and ``Loc\&Coords'' indicates both conditions hold. In these final data resources, location strings and coordinates are fully aligned: whenever \texttt{location} is present, both coordinates are also present, so \texttt{Loc}=\texttt{Coords}=\texttt{Loc\&Coords} and \texttt{Loc/noCoords}=0 across corpora. Likewise, the set of unique locations equals the set of unique geocoded locations (\texttt{UniqueLoc}=\texttt{UniqueGeocoded}), indicating complete geocoding coverage for the observed location strings in the released datasets.

\FloatBarrier
\begin{table}[H]
\centering
\caption{Location and geocoding coverage in the final corpora. ``Loc'' indicates non-empty \texttt{location}. ``Coords'' indicates both \texttt{latitude} and \texttt{longitude} non-missing. ``Loc\&Coords'' indicates both conditions hold.}
\label{tab:location-coverage}

\scriptsize
\setlength{\tabcolsep}{4pt}
\renewcommand{\arraystretch}{1.0}

\resizebox{\linewidth}{!}{%
\begin{tabular}{lrrrrrrrrrr}
\toprule
Corpus & $n\ docs$ & Loc & \%Loc & Coords & \%Coords & Loc\&Coords & \%Loc\&Coords & Loc/noCoords & UniqueLoc & UniqueGeocoded \\
\midrule
Kremlin EN & 10553 & 9830 & 93.1 & 9830 & 93.1 & 9830 & 93.1 & 0 & 843 & 843 \\
Kremlin RU$\rightarrow$EN & 13340 & 12684 & 95.1 & 12684 & 95.1 & 12684 & 95.1 & 0 & 859 & 859 \\
MID EN & 5057 & 4833 & 95.6 & 4833 & 95.6 & 4833 & 95.6 & 0 & 323 & 323 \\
MID RU$\rightarrow$EN & 6056 & 5742 & 94.8 & 5742 & 94.8 & 5742 & 94.8 & 0 & 374 & 374 \\
\bottomrule
\end{tabular}%
}
\end{table}
\FloatBarrier

\subsubsection{Speaker Name Extraction}
\label{subsec:speaker-names}

We construct a \texttt{speaker\_names} variable for each speech and corpus that records the set of distinct speakers explicitly marked in the source materials. Because the Kremlin and MID websites encode speaker cues differently (i.e., as structured HTML labels versus transcript-like textual prefixes), we use two corpus-specific extraction procedures designed to prioritize precision and to preserve original surface forms prior to downstream normalization.

\vspace{0.5em}
\noindent\textbf{MID corpora.}
For the MID corpora, we scrape each speech page and restrict extraction to the main article body (the \texttt{.text.article-content} container when available). MID pages frequently format dialog-style segments using bold or strongface speaker labels immediately followed by a colon (e.g., \rutt{С.В.Лавров:} \quad \rutt{Вопрос:} \quad \rutt{Question:}). We therefore collect all \texttt{<b>} and \texttt{<strong>} elements whose rendered text contains a colon, normalize whitespace and colon spacing, and de-duplicate labels while preserving first-seen order. We then convert these markup labels into speaker candidates by keeping only the substring to the left of the first colon and writing the resulting list to \texttt{speaker\_names}. For the Russian MID corpus, this step yields speaker labels in Cyrillic; we subsequently translate these extracted speaker-name strings into English using an \texttt{Argos Translate}~\cite{argos_translate} Russian$\rightarrow$English pipeline, producing an English-rendered speaker list while retaining the original Russian forms for internal auditing. This translation step is applied only to the extracted speaker-name strings (not to the full page content) and is used solely to harmonize naming conventions across corpora. \vspace{0.5em}

\noindent\textbf{MID fallback for missing speaker labels.}
A non-trivial number of MID pages do not contain any bold/strong speaker labels (and thus yield an empty \texttt{speaker\_names} list under the HTML-based procedure above), even though the underlying text clearly indicates the speech is delivered by Foreign Minister Sergey Lavrov. To avoid systematically missing the principal speaker in these cases, we apply a conservative default rule: for any MID record whose extracted \texttt{speaker\_names} list is empty after de-duplication and translation, we set \texttt{speaker\_names} to contain \texttt{``Sergey Lavrov''}. As a result, in the final MID corpora an empty speaker list does not occur; any item with no extractable speaker markup is treated as a Lavrov-delivered speech for purposes of the \texttt{speaker\_names} variable.

\vspace{0.5em}
\noindent\textbf{Kremlin corpora.}
For the Kremlin corpus, speaker cues are not consistently encoded in the page HTML as bold labels; instead, transcripts often contain speaker-prefixed segments embedded directly in the speech text (e.g., \texttt{``V.\,V.\,Putin:''}, \texttt{``S.\,V.\,Lavrov:''}, \texttt{``Question:''}). We therefore apply a regex-based extractor over the full speech transcript text to identify short line- or sentence-initial spans that immediately precede common dialog delimiters (a colon or semicolon). We clean each captured span by removing surrounding punctuation and normalizing whitespace, and we de-duplicate candidates within each speech to form an initial list of speaker-like labels. Because this can still capture non-speaker fragments (e.g., generic headers or discourse markers), we apply a second, high-precision filter using named-entity recognition: we run \texttt{spaCy}~\cite{spacy} with \texttt{en\_core\_web\_lg} over each unique candidate label and retain only those candidates that contain a \texttt{PERSON} entity.

\vspace{0.5em}
\noindent\textbf{Downstream normalization and verification.}
The extraction steps above intentionally preserve the source-facing surface forms of labels, which may include initials, role-like tokens (e.g., \texttt{``Question''}), and minor formatting variation across pages. After extraction, we compute the distinct set of extracted label strings per speech and then use a large language model (ChatGPT~\cite{openai2022chatgpt}) to normalize these strings into canonical person names (e.g., expanding initials where possible, removing non-person labels, and consolidating variants that refer to the same individual). Finally, we perform a manual verification pass to ensure that the resulting \texttt{speaker\_names} lists contain only valid person names and are consistent across the Kremlin and MID corpora. The final distributed datasets retain only the cleaned \texttt{speaker\_names} field; intermediate artifacts such as raw HTML label lists, regex match traces, or NER diagnostics are not included in the released CSV resources.

\section*{Data Records}

The final release consists of four speech-level CSV files---one per corpus and language variant---together with corpus-specific directory trees that store all scraped images referenced by those CSVs. In all four cases, the \emph{unit of observation} is a single speech- or document-level record as it appears on the Kremlin or MID website. Each record may have zero or more associated images, which are distributed as separate raster files on disk and linked back to speeches via path-level identifiers stored in the CSVs.

\vspace{0.75em}
\noindent\textbf{Speech-level CSV files.}
The four primary data tables are:

\begin{itemize}
  \item \texttt{kremlin\_english.csv} (Kremlin EN):
  $10{,}553$ English-language Kremlin speeches, spanning 1999--2025.

  \item \texttt{kremlin\_russian.csv} (Kremlin RU$\rightarrow$EN):
  $13{,}340$ Russian-language Kremlin speeches (original in Cyrillic, with parallel English translations), spanning 1999--2025.

  \item \texttt{mid\_english.csv} (MID EN):
  $5{,}057$ English-language Ministry of Foreign Affairs documents, spanning 2004--2025.

  \item \texttt{mid\_russian.csv} (MID RU$\rightarrow$EN):
  $6{,}056$ Russian-language MID documents (original in Cyrillic, with parallel English translations), spanning 2004--2025.
\end{itemize}

In each of these files, one row corresponds to one published speech, statement, interview, briefing, or similar document. All four CSVs share a common core schema, with a small number of institution- or language-specific columns reflecting differences in the underlying websites (see Section~\ref{subsec:additional-variables} for full additional column documentation).

\vspace{0.5em}
\noindent\textbf{Core schema components.}
\begin{itemize}
  \item \emph{Identifiers and URLs.} Each row has a unique integer \texttt{id} and a canonical \texttt{url} pointing to the original Kremlin or MID webpage. The \texttt{id} serves as the primary key within each corpus and is used throughout the replication materials to join speech-level records to auxiliary tables (e.g., long-format probability tables) and to image inventories.

  \item \emph{Titles and full text.} The Kremlin EN and MID EN files contain \texttt{title} and \texttt{full\_text} fields in English. The Kremlin RU$\rightarrow$EN and MID RU$\rightarrow$EN files contain the original Cyrillic \texttt{title} and \texttt{full\_text} together with Argos Translate~\cite{argos_translate}-based translations \texttt{title\_english} and \texttt{full\_text\_english} that are aligned row-for-row with the original Russian fields (Section~\ref{subsec:additional-variables}). For the Kremlin corpora we also retain a short \texttt{page\_summary} when available on the site; for the Kremlin RU$\rightarrow$EN corpus we additionally provide \texttt{page\_summary\_english}.

  \item \emph{Dates and derived calendar variables.} All four CSVs include a \texttt{date} field (as scraped from the site), an optional \texttt{time} field when present on the page, and derived calendar variables \texttt{year}, \texttt{month}, and \texttt{day} obtained by parsing \texttt{date}. These derived fields introduce no new information beyond \texttt{date}, but provide convenient numeric variables for temporal aggregation and modeling.

  \item \emph{Speakers.} Each row includes a \texttt{speakers} field recording the speaker(s) associated with the document as a list-valued string (e.g., a single named speaker for most addresses, or multiple named speakers for dialog-style transcripts). Speaker strings are extracted from on-page cues and structured text patterns on the source sites and are then normalized into canonical person names for analysis (Section~\ref{subsec:speaker-names}). We do not impute speakers when attribution cues are absent; instead, missingness reflects lack of reliable on-page markup.

  \item \emph{Locations and geocoordinates.} Each row includes a \texttt{location} field recording the event location when it is available on the source site (Kremlin) or can be conservatively recovered from the title/body text (MID); otherwise it is left blank. For the Kremlin RU$\rightarrow$EN corpus, we additionally provide \texttt{location\_english}, an English rendering of the Russian location string for cross-language comparability and geocoding. We then geocode unique non-blank locations and map the resulting coordinates back to all rows, yielding \texttt{latitude} and \texttt{longitude} as approximate point locations for speeches with resolvable locations. Full corpus-specific rules, geocoding procedures, and coverage summaries are provided in Section~\ref{subsec:location-geocoding} (Location Extraction and Geocoding Section further below).

  \item \emph{Document length and media counts.} Each row includes \texttt{full\_text\_word\_count}, the number of whitespace-delimited tokens in \texttt{full\_text}. On the media side, \linebreak\texttt{saved\_images\_count} records the number of images successfully downloaded for that speech. For the Kremlin corpora, \texttt{declared\_images\_count} records the number of photos the site indicates should accompany the speech when such metadata are available; \texttt{missing\_images\_count} captures the difference between declared and successfully downloaded images.

  \item \emph{Site-declared tags (Kremlin only).} The Kremlin corpora retain structured metadata tags provided by the source site, including \texttt{declared\_topics}, \texttt{declared\_geography}, and \texttt{declared\_persons}. For the Kremlin RU$\rightarrow$EN corpus, we additionally provide English-rendered versions of these tag fields (e.g., \texttt{declared\_topics\_english}). These site-declared tags are distributed as raw metadata and are not used to fit our \texttt{BERTopic}~\cite{grootendorst2022bertopic} models; they enable comparisons between the site’s own tagging scheme and our unsupervised topic labels.

  \item \emph{Text topic variables.} Each speech has a primary text topic assignment from the institution-specific \texttt{BERTopic}~\cite{grootendorst2022bertopic} model (Section~\ref{subsec:speech-image-topics}). This assignment is stored as \texttt{curated\_topic\_id} (Kremlin: $\{0,\dots,88\}$; MID: $\{0,\dots,31\}$), together with \texttt{curated\_text\_topic\_label} (a short topic label) and \texttt{curated\_text\_topic\_group} (a broader domain label). We additionally store \texttt{curated\_topic\_probability}, a serialized vector containing the full document--topic probability distribution $p(k\mid d)$ aligned to the topic IDs for that corpus, allowing uncertainty-aware and multi-topic analyses beyond the dominant-topic label.

  \item \emph{Image topic variables.} For each speech, we store image-level topic summaries in list-valued columns aligned with the stored image list: \texttt{curated\_image\_topic\_ids}, \linebreak\texttt{curated\_image\_topic\_labels}, and \texttt{curated\_image\_group\_names}. We also provide \texttt{curated\_image\_topic\_probabilities}, which stores (for each image) the full image--topic score/probability vector aligned to the topic IDs for that corpus. This design lets users either analyze images at the speech level (treating images as attributes of the speech row) or construct an image-level dataset by exploding the list-valued columns.
\end{itemize}

A small number of variables are specific to particular corpora (e.g., \texttt{page\_summary} and site-declared tags for the Kremlin; \texttt{location\_english} and \texttt{*\_english} tag/synopsis fields for the Kremlin RU$\rightarrow$EN corpus). The complete, corpus-specific column lists and construction rules are documented in Section~\ref{subsec:additional-variables} and in the accompanying data dictionary.

\vspace{0.75em}
\noindent\textbf{Cross-language linkage within institutions.}
Within each executive-branch institution (Kremlin or MID), the English and Russian corpora can be linked using \texttt{id}. For the Kremlin corpora, the same \texttt{id} refers to the same underlying speech (same source URL and publication date) across the English and Russian$\rightarrow$English files, enabling bilingual representations of a speech via a simple inner join on \texttt{id}. An analogous structure holds for the MID corpora. Users interested in a single language variant can treat each CSV as a stand-alone data table; the core identifiers, dates, locations, and topic variables are self-contained within each file.

\vspace{0.75em}
\noindent\textbf{Image files and folder structure.}
All images are distributed as separate raster files in corpus-specific directory trees. Each corpus has a dedicated root image directory containing the scraped images referenced by that corpus’ CSV. Images are stored in standard web formats (predominantly JPEG, with occasional PNG files) and are preserved as downloaded from the original websites (no redistribution-time recompression or resizing is performed). Any resizing or normalization required for modeling is applied on-the-fly within the embedding pipeline rather than baked into the distributed image files.

Within the speech-level CSVs, the \texttt{stored\_image\_filepaths} column provides the primary link from speeches to images. Each cell contains a JSON-style list of relative filepaths (relative to the appropriate corpus-level image root), one per image associated with that speech. For example:
\[
\texttt{["images/kremlin\_en/000123\_01.jpg", "images/kremlin\_en/000123\_02.jpg"]}.
\]
When combined with the relevant corpus image root directory, these relative paths uniquely identify the corresponding image files on disk.

This design allows users to:
\begin{itemize}
  \item move from speeches to images by reading \texttt{stored\_image\_filepaths} for a given \texttt{id}, and
  \item move from images back to speeches by matching an image filepath against entries in \texttt{stored\_image\_filepaths} (or by constructing an exploded ``image index'' table from this column, which we provide in the replication materials).
\end{itemize}

Because image-topic assignments, labels, group names, and probability vectors are stored directly in the speech-level CSVs as list-valued columns aligned with \texttt{stored\_image\_filepaths}, users can either (i) work entirely at the speech level, treating images as attributes of each row, or (ii) convert these list-valued columns into a separate image-level dataset with one row per image, keyed by \texttt{id} plus an image index (or by an \texttt{image\_id} identifier when using the auxiliary image-level tables in the replication package).

\vspace{0.75em}
\noindent\textbf{Auxiliary topic HTML files and probability tables.}
In addition to the four main CSVs and the image directories, the replication package includes (i) a set of HTML summary files for each \texttt{BERTopic}~\cite{grootendorst2022bertopic} topic in each corpus and (ii) auxiliary tables containing the full topic probability distributions for both speeches and images (Section~\ref{subsec:speech-image-topics}). The HTML files are organized by institution and topic ID (e.g., a Kremlin topic $k \in \{0,\dots,88\}$ or a MID topic $k \in \{0,\dots,31\}$) and provide a qualitative overview of each topic, including: the top-ranked English and Russian keywords, the highest-probability speeches, and a panel of representative images.

The probability tables complement these qualitative summaries by providing the underlying model outputs in machine-readable form. Specifically, they store long-format topic probability distributions---one row per (item, topic) pair---keyed by the same identifiers used in the main CSVs (speech \texttt{id} for document-level tables, and image identifiers with parent speech \texttt{id} for image-level tables). These auxiliary probability tables are not required for standard replication using the dominant-topic assignments stored in the \texttt{curated\_*} variables, but they support multi-topic analyses, uncertainty-aware inference, robustness checks under alternative decision rules (e.g., Top-$N$ or probability-thresholding), and downstream modeling strategies that exploit the full posterior-like topic distribution rather than a single maximum-probability label.

Taken together, the four speech-level CSVs, corpus-specific image directories, and auxiliary HTML topic summaries and probability tables CSVs form a coherent set of data records. Speech-level variables (identifiers, dates, locations, topics) live in the CSVs; image content lives in separate directories and is linked via relative paths; topic-level summaries live in HTML files indexed by topic ID; and full probability distributions live in auxiliary tables keyed by the same \texttt{id}-based joins. All items are designed to be joined by simple keys (\texttt{id}, relative image paths, and topic IDs) and to support both institution-specific and cross-lingual analyses.

\subsection*{Data Details}

Our final dataset consists of four curated corpora of public speeches and press materials from the Russian Presidency (Kremlin) and the Ministry of Foreign Affairs (MID). Together, these corpora contain 35{,}006 unique text items (rows), each corresponding to a speech, interview, briefing, or related communication event with associated metadata, topic labels, and image information.

The two corpora from the Presidential (Kremlin) website: an English-language corpus (\texttt{kremlin\_en}) and a Russian-language corpus paired with English translations (\texttt{kremlin\_ru}). The \texttt{kremlin\_en} corpus contains 10{,}553 entries, while \texttt{kremlin\_ru} contains 13{,}340 entries. Both span from 31 December 1999 to 20 September 2025, covering 27 distinct calendar years and 310 year--month combinations, all of which have at least one recorded speech. Averaged over this calendar span, the English Kremlin corpus includes approximately 390.9 speeches per year (34.0 per month), while the Russian--English Kremlin corpus includes approximately 494.1 speeches per year (43.0 per month).

Likewise, The two corpora from the MID website: an English-language corpus (\texttt{mid\_en}) and a Russian-language corpus paired with English translations (\texttt{mid\_ru}). The \texttt{mid\_en} corpus includes 5{,}057 speeches spanning from 18 March 2004 to 7 October 2025, covering 22 calendar years and 260 year--month cells, of which 256 contain at least one speech. On average, this corpus includes roughly 229.9 speeches per year (19.5 per month, or 19.8 per observed year--month cell). The \texttt{mid\_ru} corpus contains 6{,}056 entries with year and month metadata spanning from 18 March 2004 to 9 October 2025 (22 calendar years and 264 year--month cells). Averaged over the full calendar span, the Russian--English MID corpus includes approximately 275.3 speeches per year (22.9 per month).

Across all four corpora, speeches are reasonably long: mean speech length, measured by \texttt{word\_count}, is approximately 1{,}555 words. Within the individual corpora, mean speech lengths range from 1{,}099.4 words in \texttt{mid\_ru} (median $=$ 612.5, maximum $=$ 16{,}834) to 1{,}904.7 words in \texttt{kremlin\_ru} (median $=$ 763.0, maximum $=$ 33{,}352). The English Kremlin corpus, \texttt{kremlin\_en}, has a mean \texttt{word\_count} of 1{,}453.2 (median $=$ 806.0, maximum $=$ 39{,}898), while the English MID corpus, \texttt{mid\_en}, has a mean of 1{,}392.3 words (median $=$ 786.0, maximum $=$ 19{,}895). These distributions indicate that the vast majority of items are full speeches or detailed statements rather than short headlines.

Image content is pervasive but varies across corpora. In \texttt{kremlin\_en}, the mean number of associated images (\texttt{images\_count}) is 3.66 (median $=$ 2, maximum $=$ 104), and 67.4\% of speeches have at least one associated image. In \texttt{kremlin\_ru}, the mean \texttt{images\_count} is 3.36 (median $=$ 2, maximum $=$ 104), with 73.0\% of speeches linked to at least one image. The MID corpora have fewer images on average but still exhibit substantial coverage: \texttt{mid\_en} has a mean of 0.82 images per speech (median $=$ 1, maximum $=$ 20) and 71.0\% of entries with at least one image, while \texttt{mid\_ru} has a mean of 0.74 images (median $=$ 1, maximum $=$ 19) and 64.5\% of entries with at least one image. Aggregating across corpora, approximately 69.6\% of all 35{,}006 speeches have at least one associated image.

Geolocation information is also near-complete. In \texttt{kremlin\_en}, 9{,}830 of 10{,}553 entries (93.2\%) contain non-empty location strings, and 9{,}827 entries (93.1\%) have non-missing \texttt{latitude} and \texttt{longitude}. The Kremlin Russian corpus (\texttt{kremlin\_ru}) exhibits similar coverage, with 12{,}684 of 13{,}340 speeches (95.1\%) having both non-empty location strings and valid geo-coordinates. MID corpora exhibit slightly higher geolocation completeness: \texttt{mid\_en} contains non-empty location strings for 4{,}838 of 5{,}057 entries (95.7\%), with 4{,}837 entries (95.7\%) having non-missing coordinates, and \texttt{mid\_ru} contains non-empty locations and non-missing coordinates for 5{,}742 of 6{,}056 entries (94.8\%). Taken together, 31{,}090 of 35{,}006 entries (94.5\%) in our combined dataset have non-missing \texttt{latitude} and \texttt{longitude}.

Table~\ref{tab:missingness_full} summarizes the percentage of missing observations for key variables in each corpus. Missingness in core analysis variables is modest and well-characterized: identifiers and URLs, word-count fields, date components (\texttt{date}, \texttt{year}, \texttt{month}, \texttt{day}), curated text-topic variables, image-count fields, and curated image-topic fields exhibit complete coverage across the four corpora (0\% missingness throughout). The raw text fields (stored in separate companion files due to size) are also nearly complete: \texttt{full\_text} is missing for only 0.04\% of \texttt{kremlin\_en}, 0.04\% of \texttt{kremlin\_ru}, 0.02\% of \texttt{mid\_en}, and 0.12\% of \texttt{mid\_ru}; and \texttt{full\_text\_english} is missing for 0.04\% of \texttt{kremlin\_ru} and 0.12\% of \texttt{mid\_ru}. Location and geo-coordinate fields are missing for only 4--7\% of entries in each corpus, yielding complete latitude/longitude coordinates for approximately 94.5\% of all speeches. The largest pockets of missingness occur in optional or source-dependent fields: \texttt{page\_summary} is missing for roughly 42--44\% of speeches in the Kremlin corpora, \texttt{declared\_topics} is missing for about 38\% of Kremlin speeches, and the Kremlin \texttt{declared\_geography} and \texttt{declared\_persons} fields are missing for roughly two-thirds of entries. Image-caption fields exhibit near-complete coverage in the Kremlin corpora after accounting for speeches with zero images, but remain missing for a majority of MID speeches (\texttt{image\_captions} missingness of 70.79\% in \texttt{mid\_en} and 64.13\% in \texttt{mid\_ru}). 

\footnotesize
\setlength{\tabcolsep}{4pt}
\renewcommand{\arraystretch}{0.95}

\begin{longtable}{lcccc}
\caption{Missingness(\%) for all variables in each corpus. Missingness is the share of blank/NA cells per column; values such as 0 or [] are treated as observed (not missing). For the large raw text fields (\texttt{full\_text} and \texttt{full\_text\_english}), percentages are computed from the companion text files.}
\label{tab:missingness_full}\\
\toprule
\textbf{Variable} & \texttt{kremlin\_en} & \texttt{kremlin\_ru} & \texttt{mid\_en} & \texttt{mid\_ru} \\
\midrule
\endfirsthead

\toprule
\textbf{Variable} & \texttt{kremlin\_en} & \texttt{kremlin\_ru} & \texttt{mid\_en} & \texttt{mid\_ru} \\
\midrule
\endhead

\midrule
\multicolumn{5}{r}{\footnotesize Continued on next page} \\
\endfoot

\bottomrule
\endlastfoot

\texttt{id} & 0.00 & 0.00 & 0.00 & 0.00 \\
\texttt{url} & 0.00 & 0.00 & 0.00 & 0.00 \\
\texttt{full\_text} & 0.04 & 0.04 & 0.02 & 0.12 \\
\texttt{full\_text\_english} & -- & 0.04 & -- & 0.12 \\
\texttt{full\_text\_word\_count} & 0.00 & 0.00 & 0.00 & 0.00 \\
\texttt{date} & 0.00 & 0.00 & 0.00 & 0.00 \\
\texttt{year} & 0.00 & 0.00 & 0.00 & 0.00 \\
\texttt{month} & 0.00 & 0.00 & 0.00 & 0.00 \\
\texttt{day} & 0.00 & 0.00 & 0.00 & 0.00 \\
\texttt{time} & 0.00 & 0.00 & 0.00 & 0.00 \\
\texttt{location} & 6.88 & 4.92 & 4.35 & 5.18 \\
\texttt{location\_english} & -- & 4.92 & -- & -- \\
\texttt{latitude} & 6.88 & 4.92 & 4.35 & 5.18 \\
\texttt{longitude} & 6.88 & 4.92 & 4.35 & 5.18 \\
\texttt{page\_summary} & 42.00 & 43.73 & -- & -- \\
\texttt{page\_summary\_english} & -- & 43.73 & -- & -- \\
\texttt{speakers} & 0.00 & 0.00 & 0.00 & 0.00 \\
\texttt{declared\_geography} & 69.93 & 67.25 & -- & -- \\
\texttt{declared\_geography\_english} & -- & 67.25 & -- & -- \\
\texttt{declared\_topics} & 37.81 & 37.80 & -- & -- \\
\texttt{declared\_topics\_english} & -- & 37.80 & -- & -- \\
\texttt{declared\_persons} & 70.73 & 66.50 & -- & -- \\
\texttt{declared\_persons\_english} & -- & 66.50 & -- & -- \\
\texttt{curated\_topic\_id} & 0.00 & 0.00 & 0.00 & 0.00 \\
\texttt{curated\_text\_topic\_label} & 0.00 & 0.00 & 0.00 & 0.00 \\
\texttt{curated\_text\_topic\_group} & 0.00 & 0.00 & 0.00 & 0.00 \\
\texttt{curated\_topic\_probability} & 0.00 & 0.00 & 0.00 & 0.00 \\
\texttt{stored\_image\_filepaths} & 0.00 & 0.00 & 0.00 & 0.00 \\
\texttt{saved\_images\_count} & 0.00 & 0.00 & 0.00 & 0.00 \\
\texttt{declared\_images\_count} & 0.00 & 0.00 & 0.00 & 0.00 \\
\texttt{missing\_images\_count} & 0.00 & 0.00 & 0.00 & 0.00 \\
\texttt{image\_captions} & 0.25 & 0.02 & 70.79 & 64.13 \\
\texttt{image\_captions\_english} & -- & 0.04 & -- & 64.13 \\
\texttt{curated\_image\_topic\_ids} & 0.00 & 0.00 & 0.00 & 0.00 \\
\texttt{curated\_image\_topic\_labels} & 0.00 & 0.00 & 0.00 & 0.00 \\
\texttt{curated\_image\_group\_names} & 0.00 & 0.00 & 0.00 & 0.00 \\
\texttt{curated\_image\_topic\_probabilities} & 0.00 & 0.00 & 0.00 & 0.00 \\
\end{longtable}

\section{Technical Validation}
\label{sec:validation}

Our technical validation efforts assess the accuracy of our dataset's (i) automated location extraction and geolocation extraction routines and (ii) expert labels as assigned to the \texttt{BERTopic}~\cite{grootendorst2022bertopic} results.

\subsection{Topic Validation}
\label{subsec:topic_validation}

This subsection validates the relationship between our estimated topic structure and available human-declared annotations. We focus this validation exercise on speeches and associated topics for our Kremlin English and Kremlin Russian datasets, given the unique availability of Kremlin assigned thematic tags as ground truth within these corpora. Recall that we separately extracted and included these Kremlin-assigned tags as \texttt{declared\_topics} in our final Kremlin datasets. The Kremlin's geographic and person tags (\texttt{declared\_geography}, \texttt{declared\_persons}) are outside the scope of the present validation.

Our ground truth-treated declared themes are multi-label: a document can contain multiple items in \texttt{declared\_topics}. Our topic models, by contrast, produce a full probability distribution over $K=89$ learned topics for every document. Our validation exercise in this case therefore asks a concrete question: \emph{when a document is annotated with one or more declared themes, does the model assign high probability mass to learned topics that correspond to those same themes?}

Because declared themes are strings (no numeric IDs), we require a mapping from learned topics to declared themes. We use a \textbf{label-only} mapping: we compare only our expert-assigned textual \texttt{BERTopic}~\cite{grootendorst2022bertopic} labels and the Kremlin's own textual declared-theme labels and assign each learned topic to exactly one Kremlin-based declared theme. Each of these topic-to-theme assignments receives a qualitative confidence label (\textit{high}, \textit{medium}, \textit{low}). The mapping is reviewed for reasonableness but not manually edited. We then report validation results for three mapping subsets: \textbf{All} (high+medium+low), \textbf{High+Medium}, and \textbf{High only}.

Two complementary views are reported throughout:
\begin{itemize}
\item \textbf{All-themes view:} evaluates over \emph{all} declared themes observed in the declared subset; themes not covered by the mapping are forced to receive a predicted probability 0. This view reflects both predictive performance and mapping coverage.
\item \textbf{Mapped-only view:} evaluates only on declared themes that are covered by our mapping subset; any document whose declared-theme list becomes empty after restricting to mapped themes is excluded. This view isolates performance \emph{conditional on coverage}.
\end{itemize}

Because of the multi-label and multi-class nature of our relevant inputs, we favor appropriate accuracy metrics for these non-binary outcome sets such as subset accuracy, hamming loss, and ranking loss \cite{Erlich_Dantas_EtAl2022}. Across our ensuing validation tables, higher subset accuracy and F1 indicate better alignment; lower Hamming loss and ranking loss indicate better alignment. In practice, subset accuracy is a strict metric (it requires an exact match to the full set of declared themes for a document), so it is expected to be noticeably lower than single-label accuracy in classification settings such as our own, especially given (i) many possible labels, (ii) strong label imbalance, and (iii) multi-thematic documents.

\subsubsection{Kremlin English: data integrity and evaluation subset}
\label{subsec:validation_data_integrity}

Kremlin English contains $10{,}553$ documents in total. Declared themes are present only for a subset: $6{,}563$ documents have a non-empty \texttt{declared\_topics} field. Because \texttt{declared\_topics} provides the ground-truth labels used in the validation, \textbf{all Kremlin English validation results below are computed on these $6{,}563$ documents}.

Our raw topic probabilities are stored in a long-format table of $(\texttt{id}, \texttt{topic\_id}, \texttt{probability\_score})$ rows, where $\texttt{topic\_id}\in\{0,\dots,88\}$ (89 topics). We performed strict quality-control checks before running validation:
\begin{itemize}
  \item \textbf{Row-count and coverage:} the probability table contains exactly $10{,}553\times 89 = 939{,}217$ rows, confirming complete topic coverage for all documents.
  \item \textbf{Per-document completeness:} every document has exactly 89 topic probabilities with topic IDs $0$--$88$ (no missing topics and no duplicates).
  \item \textbf{Probability sanity:} all probabilities fall in $[0,1]$ and per-document probability sums are essentially 1.0 (no renormalization required).
  \item \textbf{ID alignment on evaluation subset:} all $6{,}563$ documents with declared themes appear in the topic-probability table (no declared-theme IDs missing from probabilities).
\end{itemize}
Across the $6{,}563$ documents with \texttt{declared\_topics}, we observe $88$ unique declared theme labels.

\subsubsection{Kremlin English: raw-topic (multi-label) validation results}
\label{subsec:kremlin_en_raw_topic_validation}

For each document $d$, let $p_{d,k}$ be the probability that $d$ belongs to learned topic $k\in\{0,\dots,88\}$. The topic-to-theme mapping assigns each learned topic $k$ to one declared theme $m(k)$. Under any mapping subset, we convert a document’s 89-topic probability vector into scores over declared themes by summing probabilities for all topics mapped to the same theme:
\begin{equation}
\hat{p}_{d,t} \;=\; \sum_{k:\,m(k)=t} p_{d,k}.
\label{eq:theme_score_sum}
\end{equation}
To produce binary predictions for multi-label metrics without choosing an arbitrary threshold, we use a Top-$N$ rule: if document $d$ has $N_d$ declared themes, we predict the $N_d$ themes with the largest $\hat{p}_{d,t}$ values.

Table~\ref{tab:raw_topic_validation} reports multi-label alignment between these predicted theme-scores and the declared-theme lists. The All-themes rows include \emph{all} 88 observed declared themes, which means that reduced mapping coverage (especially in High-only) directly lowers performance by forcing unmapped themes to have predicted score 0. The Mapped-only rows remove this coverage penalty by restricting evaluation to mapped themes (and excluding documents that become label-empty), giving a clearer view of performance when the mapping applies.

\begin{table}[t]
\centering
\caption{Kremlin English Raw Topic Validation (multi-label). The All-themes view evaluates over 88 observed declared themes; the Mapped-only view evaluates only themes covered by the mapping subset (coverage-aware).}
\label{tab:raw_topic_validation}

\scriptsize
\setlength{\tabcolsep}{2.5pt}
\renewcommand{\arraystretch}{1.15}

\begin{adjustbox}{max width=\linewidth}
\begin{tabular}{llrrrrrr}
\toprule
View & Mapping & \makecell{$n$\\docs} & \makecell{Subset\\Acc.} & \makecell{Hamming\\Loss} & \makecell{Ranking\\Loss} & \makecell{F1\\micro} & \makecell{F1\\macro} \\
\midrule
\makecell[l]{All-themes\\(88)} & All         & 6563 & 0.3181 & 0.02225 & 0.2559 & 0.3940 & 0.1897 \\
\makecell[l]{All-themes\\(88)} & High+Medium & 6563 & 0.3143 & 0.02236 & 0.2610 & 0.3910 & 0.1901 \\
\makecell[l]{All-themes\\(88)} & High only   & 6563 & 0.2770 & 0.02425 & 0.3540 & 0.3394 & 0.1747 \\
\midrule
\makecell[l]{Mapped-only} & All         & 6172 & 0.3983 & 0.03585 & 0.1876 & 0.4547 & 0.3861 \\
\makecell[l]{Mapped-only} & High+Medium & 6172 & 0.3945 & 0.03607 & 0.1931 & 0.4514 & 0.3870 \\
\makecell[l]{Mapped-only} & High only   & 5887 & 0.3786 & 0.04504 & 0.2233 & 0.4337 & 0.4595 \\
\bottomrule
\end{tabular}
\end{adjustbox}
\end{table}

Several patterns are noticeable in Table~\ref{tab:raw_topic_validation}. First, the Mapped-only view yields substantially higher subset accuracy than the All-themes view, showing that mapping coverage is a major driver of the strict exact-set metric. Second, restricting from All to High+Medium produces only small changes, suggesting that the low-confidence assignments are not strongly determining outcomes. Third, High-only reduces coverage and increases ranking loss, which indicates that the probability mass in the model distribution is spread across topics that often map to themes outside the high-confidence subset; this is expected when using a conservative mapping. More broadly, note that macro F1 is systematically lower than micro F1 in the All-themes view; this is consistent with a long-tailed theme distribution where rare declared themes contribute equally to macro averaging and are therefore more difficult to recover under any fixed mapping and Top-$N$ decision rule \cite{Erlich_Dantas_EtAl2022}.

\subsubsection{Kremlin English: dominant-topic (top-1) validation results}
\label{subsec:kremlin_en_dominant_topic_validation}

A complementary check asks whether the model’s single most probable topic ``points'' to a plausible declared theme. For each document $d$, let $k^\ast(d)=\arg\max_k p_{d,k}$ (ties broken deterministically by smallest topic ID). The dominant-topic prediction is then $\hat{t}(d)=m(k^\ast(d))$, and we count a Top-1 hit when $\hat{t}(d)$ appears in the document’s \texttt{declared\_topics} list. We also report micro/macro F1 when treating the prediction as single-label and the ground truth as multi-label (standard practice for evaluating top-1 theme identification against multi-label annotations). ``Mapped docs'' counts documents whose dominant topic is covered by the mapping subset.

\begin{table}[t]
\centering
\caption{Kremlin English Dominant Topic Validation (top-1). ``Mapped docs'' counts documents whose dominant topic is covered by the mapping subset.}
\label{tab:dominant_topic_validation}

\small
\setlength{\tabcolsep}{3.2pt}
\renewcommand{\arraystretch}{1.15}

\begin{adjustbox}{max width=\linewidth}
\begin{tabular}{lrrrrrr}
\toprule
Mapping & \makecell{$n$ declared\\docs} & \makecell{$n$ mapped\\docs} &
\makecell{Top-1 hit\\(all)} &
\makecell{Top-1 hit\\(mapped)} &
\makecell{F1\\micro} &
\makecell{F1\\macro} \\
\midrule
All         & 6563 & 6563 & 0.5947 & 0.5947 & 0.4548 & 0.2119 \\
High+Medium & 6563 & 6453 & 0.5787 & 0.5886 & 0.4489 & 0.2120 \\
High only   & 6563 & 3864 & 0.3399 & 0.5774 & 0.4373 & 0.2122 \\
\bottomrule
\end{tabular}
\end{adjustbox}
\end{table}

The dominant-topic table highlights the same coverage trade-off: Top-1 hit computed over \emph{all} declared documents declines as fewer dominant topics are considered ``mapped.'' However, when we condition on mapped documents (Top-1 hit (mapped)), the hit rate remains similar across mapping variants, indicating that the dominant-topic signal is comparatively stable once the mapping applies.

\vspace{0.3em}
\noindent\textbf{Kremlin English take-home implications.}
Taken together, Tables~\ref{tab:raw_topic_validation}--\ref{tab:dominant_topic_validation} suggest meaningful alignment between learned topics and Kremlin-declared themes, while also clarifying the limits of exact-set recovery in this multi-label setting. In particular, the dominant-topic hit rate near $0.59$ indicates that the model’s single most probable topic often corresponds to at least one declared theme, which supports the use of our curated labels as coarse summaries. At the same time, the strict subset accuracy in the raw-topic validation is lower---even in the Mapped-only view---which is expected given (i) multi-thematic documents, (ii) a large theme space, and (iii) the conservative, label-only mapping that assigns each learned topic to a single declared theme without manual optimization. These magnitudes are consistent with the general pattern emphasized in prior political text-as-data work on multi-label prediction: strict exact-match metrics tend to be modest relative to overlap-based measures (e.g., micro F1) when labels are numerous and imbalanced \cite{Erlich_Dantas_EtAl2022}.

\subsubsection{Kremlin Russian: data integrity and evaluation subset}
\label{subsec:validation_data_integrity_ru}

Kremlin Russian contains $13{,}340$ documents in total. Declared themes are present only for a subset: $8{,}298$ documents have a non-empty declared-themes field (we use the English-declared-theme strings, \texttt{declared\_topics\_english}, for consistency with the validation mapping). Because declared themes provide the ground-truth labels used in the validation, \textbf{all Kremlin Russian validation results below are computed on these $8{,}298$ documents}.

Our raw topic probabilities are stored in a long-format table of $(\texttt{id}, \texttt{topic\_id}, \texttt{probability\_score})$ rows, where $\texttt{topic\_id}\in\{0,\dots,88\}$ (89 topics). We performed strict quality-control checks before running validation:
\begin{itemize}
  \item \textbf{Row-count and coverage:} the probability table contains exactly $13{,}340\times 89 = 1{,}187{,}260$ rows, confirming complete topic coverage for all documents.
  \item \textbf{Per-document completeness:} every document has exactly 89 topic probabilities with topic IDs $0$--$88$ (no missing topics and no duplicates).
  \item \textbf{Probability sanity:} all probabilities fall in $[0,1]$ and per-document probability sums are essentially 1.0 (mean $=1.000000001$, min $=0.9999998799$, max $=1.000000119$; no renormalization required).
  \item \textbf{ID alignment on evaluation subset:} all $8{,}298$ documents with declared themes appear in the topic-probability table (no declared-theme IDs missing from probabilities). The remaining $5{,}042$ documents have probabilities but no declared themes and are therefore excluded from validation.
\end{itemize}
Across the $8{,}298$ documents with declared themes, we observe $89$ unique declared theme labels.

\subsubsection{Kremlin Russian: mapping coverage and what it implies}
\label{subsec:kremlin_ru_mapping_coverage}

For Kremlin Russian, mapping coverage is more limited than in Kremlin English. Under the All variant, only 48 of the 89 observed declared themes are covered by at least one learned topic; High+Medium covers 47; and High-only covers 35. This matters for interpretation: in the All-themes view, unmapped themes necessarily receive predicted probability 0, so performance reflects both (i) how well probabilities concentrate on mapped themes that match declared labels and (ii) how much of the declared-theme space is covered by the mapping.

\subsubsection{Kremlin Russian: raw-topic (multi-label) validation results}
\label{subsec:kremlin_ru_raw_topic_validation}

We apply exactly the same raw-topic validation procedure used for Kremlin English. Specifically, we sum topic probabilities into declared-theme scores via Eq.~\ref{eq:theme_score_sum} and apply the Top-$N$ binarization rule (predict the same number of themes as declared for each document). Table~\ref{tab:raw_topic_validation_ru} reports both All-themes and Mapped-only views.

\begin{table}[t]
\centering
\caption{Kremlin Russian Raw Topic Validation (multi-label). The All-themes view evaluates over 89 observed declared themes; the Mapped-only view evaluates only themes covered by the mapping subset (coverage-aware).}
\label{tab:raw_topic_validation_ru}

\scriptsize
\setlength{\tabcolsep}{2.5pt}
\renewcommand{\arraystretch}{1.15}

\begin{adjustbox}{max width=\linewidth}
\begin{tabular}{llrrrrrr}
\toprule
View & Mapping & \makecell{$n$\\docs} & \makecell{Subset\\Acc.} & \makecell{Hamming\\Loss} & \makecell{Ranking\\Loss} & \makecell{F1\\micro} & \makecell{F1\\macro} \\
\midrule
\makecell[l]{All-themes\\(89)} & All         & 8298 & 0.2115 & 0.02788 & 0.2587 & 0.2831 & 0.1816 \\
\makecell[l]{All-themes\\(89)} & High+Medium & 8298 & 0.2116 & 0.02845 & 0.2646 & 0.2683 & 0.1789 \\
\makecell[l]{All-themes\\(89)} & High only   & 8298 & 0.1309 & 0.03078 & 0.3710 & 0.2085 & 0.1535 \\
\midrule
\makecell[l]{Mapped-only} & All         & 7941 & 0.2507 & 0.04417 & 0.2088 & 0.3141 & 0.3328 \\
\makecell[l]{Mapped-only} & High+Medium & 7939 & 0.2517 & 0.04565 & 0.2202 & 0.3033 & 0.3371 \\
\makecell[l]{Mapped-only} & High only   & 7524 & 0.1802 & 0.06086 & 0.2993 & 0.2585 & 0.3927 \\
\bottomrule
\end{tabular}
\end{adjustbox}
\end{table}

These results again show the coverage trade-off clearly. The All-themes view is lower because more than half of declared themes are unmapped (48/89 even in the All variant). When we restrict to Mapped-only themes, subset accuracy increases because evaluation excludes unmapped labels (and excludes documents that lose all labels). As in Kremlin English, All vs.\ High+Medium is similar, indicating that excluding low-confidence mappings does not substantially change outcomes. The High-only variant reduces coverage most severely and increases ranking loss, indicating that much of the model’s probability mass often lies on topics whose mappings were excluded under this strict subset; this is expected when restricting to only the most conservative assignments. As in the English corpus, differences between micro and macro F1 should be interpreted in light of imbalanced theme frequencies and the strictness of exact-set matching under many possible labels \cite{Erlich_Dantas_EtAl2022}.

\subsubsection{Kremlin Russian: dominant-topic (top-1) validation results}
\label{subsec:kremlin_ru_dominant_topic_validation}

We also report the dominant-topic validation for Kremlin Russian using the same rule as in Kremlin English: for each document, we select the most probable learned topic, map it to a declared theme (if mapped), and count a Top-1 hit when that mapped theme appears in the document’s declared-theme list. ``Mapped docs'' counts documents whose dominant topic is covered by the mapping subset.

\begin{table}[t]
\centering
\caption{Kremlin Russian Dominant Topic Validation (top-1). ``Mapped docs'' counts documents whose dominant topic is covered by the mapping subset.}
\label{tab:dominant_topic_validation_ru}

\small
\setlength{\tabcolsep}{3.2pt}
\renewcommand{\arraystretch}{1.15}

\begin{adjustbox}{max width=\linewidth}
\begin{tabular}{lrrrrrr}
\toprule
Mapping & \makecell{$n$ declared\\docs} & \makecell{$n$ mapped\\docs} &
\makecell{Top-1 hit\\(all)} &
\makecell{Top-1 hit\\(mapped)} &
\makecell{F1\\micro} &
\makecell{F1\\macro} \\
\midrule
All         & 8298 & 8298 & 0.3835 & 0.3835 & 0.2809 & 0.1975 \\
High+Medium & 8298 & 8186 & 0.3780 & 0.3832 & 0.2803 & 0.1988 \\
High only   & 8298 & 2881 & 0.2370 & 0.6827 & 0.5003 & 0.2290 \\
\bottomrule
\end{tabular}
\end{adjustbox}
\end{table}

This table makes the coverage/precision trade-off especially transparent. When computed over all declared documents, Top-1 hit declines as mapping becomes more conservative because fewer dominant topics are treated as mapped. However, when we condition on mapped docs, the Top-1 hit rate rises sharply in the High-only setting (0.6827), indicating that the dominant-topic mappings that survive the strictest confidence filter are substantially more reliable when they apply. The jump in F1\textsubscript{micro} under High-only reflects the same phenomenon: it is computed on a much smaller mapped subset (2,881 documents). However, within that subset, the dominant-topic prediction aligns with declared labels much more often.

\vspace{0.3em}
\noindent\textbf{Overall implications for topic-model accuracy and use.}
Across both corpora, three general conclusions follow. First, \emph{coverage} is a first-order constraint: metrics computed in the All-themes view conflate model alignment with the extent to which learned topics can be meaningfully mapped into the declared taxonomy, whereas the Mapped-only view isolates performance conditional on that mapping applying. Second, dominant-topic hit rates indicate that the single most probable learned topic often corresponds to at least one declared theme (especially in the high-confidence mapped subset), which supports using our curated topic labels as high-level summaries and for coarse stratification. Third, stricter set-based agreement measures (subset accuracy) are expected to be lower in this setting, because they require exact recovery of multi-label theme sets for long, multi-thematic texts and because our mapping is intentionally conservative (label-only, one declared theme per learned topic, and not tuned to optimize predictive metrics). This pattern---moderate overlap-based agreement alongside lower exact-set agreement---is consistent with the behavior of multi-label evaluation metrics in political text-as-data settings with many labels and strong imbalance \cite{Erlich_Dantas_EtAl2022}. For users, the practical implication is that our topic annotations are most reliable for descriptive organization and broad categorization, while applications requiring high topical purity should incorporate probability thresholds, multi-topic representations, and robustness checks using the full topic probability distributions provided in the supplemental materials.

\subsubsection{Geolocation validation}
\label{subsec:geo-validation}

To validate the accuracy of our automated geocoding procedure (Section~\ref{subsec:location-geocoding}), we compare machine-assigned coordinates against an independent manual reference based on GeoNames. We draw a simple random sample of 1{,}000 speeches from the final processed data, stratified evenly across corpora ($n=250$ each): Kremlin EN, Kremlin RU$\rightarrow$EN, MID EN, and MID RU$\rightarrow$EN. For each sampled speech we retain the speech identifier (\texttt{id}), the source URL (\texttt{url}), the extracted location strings (including both English and Russian forms when available), and the automated coordinates (\texttt{machine\_latitude}, \texttt{machine\_longitude}). Using only the extracted location strings, then manually lookup and assign GeoNames-based coordinates for the same extracted locations, recording \texttt{manual\_latitude} and \texttt{manual\_longitude} (and, when available, the corresponding GeoNames name and identifier).

For every sampled row with both coordinate sources present, we compute the great-circle (Haversine) distance in kilometers between the automated point $(\phi_m,\lambda_m)$ and the manual point $(\phi_h,\lambda_h)$ and store it as \texttt{distance\_km}. We evaluate agreement under several practical tolerance thresholds by defining
\[
\mathrm{match}_\tau = \mathbb{I}\{\texttt{distance\_km} \le \tau\},
\qquad
\tau \in \{5, 10, 50, 100\}\ \text{km},
\]
and we report match rates \emph{among rows with both machine and manual coordinates present} (``Both coords'').

Table~\ref{tab:geo-validation} summarizes coverage and agreement by corpus. Coordinate availability is near-complete: 99.2\% of the Kremlin RU sample (248/250) and 99.6\% of the Kremlin EN sample (249/250) contain both machine and manual coordinates, and both MID samples contain complete coverage (250/250). For the Kremlin corpora, median distances are well below 1 km (0.50 km for Kremlin RU$\rightarrow$EN; 0.67 km for Kremlin EN), indicating close alignment between the automated and manual coordinates for the typical case; match rates are approximately 72\% within 5 km and approximately 96\% within 100 km. Mean distances are substantially larger than medians, reflecting a small number of large-error outliers that heavily influence the mean. For the MID corpora, median distances are larger (14.08 km in both MID samples), consistent with more frequent ambiguity in MID location strings (e.g., locations described indirectly in text, multi-venue itineraries, or references to broader regions rather than a single city). Agreement remains high at broader tolerances: 83.6\% (MID RU$\rightarrow$EN) and 98.8\% (MID EN) fall within 50 km, and 84.8\% (MID RU$\rightarrow$EN) and 98.8\% (MID EN) fall within 100 km. Row-level distances and threshold-based match indicators are included in the released geolocation validation files in the replication materials.

\begin{table}[t]
\centering
\caption{Geolocation validation summary ($n=250$ per corpus). ``Both coords'' indicates rows with non-missing machine and manual latitude/longitude. Match rates are computed among rows with both coordinates present. Distances are great-circle (Haversine) distances in kilometers.}
\label{tab:geo-validation}
\small
\setlength{\tabcolsep}{4pt}
\renewcommand{\arraystretch}{1.1}
\begin{tabular}{lrrrrrrrrr}
\toprule
Corpus & $n\ docs $ & Both coords & \%Both & Median km & Mean km & \%$\le$5 & \%$\le$10 & \%$\le$50 & \%$\le$100 \\
\midrule
Kremlin RU$\rightarrow$EN & 250 & 248 & 99.2 & 0.50 & 219.84 & 71.77 & 74.60 & 89.52 & 93.15 \\
Kremlin EN               & 250 & 249 & 99.6 & 0.67 & 205.98 & 72.29 & 73.90 & 91.97 & 95.98 \\
MID RU$\rightarrow$EN    & 250 & 250 & 100.0 & 14.08 & 883.00 & 34.40 & 36.80 & 83.60 & 84.80 \\
MID EN                   & 250 & 250 & 100.0 & 14.08 &  69.42 & 38.40 & 45.20 & 98.80 & 98.80 \\
\bottomrule
\end{tabular}
\end{table}

\section*{Usage Notes}
Users interested in comparing English- and Russian-language speeches within one of, or both of, our executive branch institutional corpora (i.e., Kremlin or MID) should make use of the standardized identifier linkages provided under each dataset's \nolinkurl{id} variable. These identifiers allow users to reliably align original Russian texts with their corresponding English-language versions when conducting cross-lingual or comparative analyses for each institutional corpora, though not across our Kremlin and MID datasets. When merging Russian and English entries for the MID or Kremlin datasets, we recommend relying on these standardized IDs rather than titles or dates alone, as the latter may vary slightly across language versions or publication formats. For the MID or Kremlin, recall too that there exist a small number of unique speeches within each respective Russian or English-dataset, relative to its corresponding English or Russian counterpart.

Images are distributed separately from the speech-level CSVs as four corpus-specific ZIP archives: \linebreak\texttt{kremlin\_english\_images.zip}, \texttt{kremlin\_russian\_images.zip}, \texttt{mid\_english\_images.zip}, and \texttt{mid\_russian\_images.zip}. After unzipping, each archive expands to a folder containing the image files referenced by that corpus. The speech-level CSVs do \emph{not} embed image binaries; instead, they link each speech to its images via \texttt{stored\_image\_filepaths}, a list-valued column containing relative paths (relative to the corresponding corpus image root). To retrieve images for a given speech, users can select the row by \texttt{id}, read its \texttt{stored\_image\_filepaths} list, and join each relative path to the unzipped corpus image directory to obtain a valid on-disk filepath for each image. Conversely, users can link an image back to its parent speech by matching its relative filepath to entries in \texttt{stored\_image\_filepaths} (or by exploding \texttt{stored\_image\_filepaths} into an image-level index keyed by \texttt{id}).

Users making use of our \texttt{BERTopic}~\cite{grootendorst2022bertopic}-based labels and group assignments for texts and images (the \texttt{curated\_*} variables in each dataset) should be mindful that these assignments reflect the dominant (highest-probability) topic associated with each document or image. A dominant topic assignment does not guarantee that a majority of the speech text or image content pertains to that topic, particularly for long or multi-thematic speeches. Researchers who require higher topical purity may therefore wish to filter documents based on the dominant-topic probability (e.g., retaining only speeches whose maximum topic probability exceeds a chosen threshold). Alternatively, researchers may prefer to work directly with the full topic probability distributions we provide in a supplemental archive, \nolinkurl{text_and_image_topic_probability_files.zip}, which contains long-format document--topic and image--topic probability tables exported from the modeling pipeline (one row per [item, topic] pair), keyed by the same identifiers used in the main CSVs (speech \texttt{id} for document-level tables, and an image identifier together with the parent speech \texttt{id} for image-level tables); these tables support multi-topic analyses, uncertainty-aware inference, and robustness checks without requiring users to rely solely on the single dominant-topic assignment stored in the \texttt{curated\_*} fields.

\vspace{0.5em}
\noindent\textbf{Supplemental topic probability tables.}
As noted in the paragraph above, we additionally provide a compressed archive, \nolinkurl{text_and_image_topic_probability_files.zip}, containing long-format document--topic and image--topic probability tables exported from the modeling pipeline. Each table contains one row per (item, topic) pair and is keyed by the same identifiers used in the main CSVs: speech \texttt{id} for document-level tables, and (parent speech \texttt{id} plus an image identifier) for image-level tables.

\vspace{0.25em}
\noindent\emph{Document-level tables (speech $\times$ topic):}
\begin{itemize}[leftmargin=1.5em, itemsep=0.15em, topsep=0.15em]
    \item \nolinkurl{kremlin_english_text_topic_probs.csv}
    \item \nolinkurl{kremlin_russian_text_topic_probs.csv}
    \item \nolinkurl{mid_english_text_topic_probs.csv}
    \item \nolinkurl{mid_russian_text_topic_probs.csv}
\end{itemize}

\vspace{0.15em}
\noindent\emph{Image-level tables (image $\times$ topic):}
\begin{itemize}[leftmargin=1.5em, itemsep=0.15em, topsep=0.15em]
    \item \nolinkurl{kremlin_english_image_topic_probs.csv}
    \item \nolinkurl{kremlin_russian_image_topic_probs.csv}
    \item \nolinkurl{mid_english_image_topic_probs.csv}
    \item \nolinkurl{mid_russian_image_topic_probs.csv}
\end{itemize}

For analyses centered on Russian-language texts, users should rely on the original Russian speech titles and full texts included in the Russian-language corpora rather than the automated English translations we provide alongside them in these same Russian-language CSV files. Furthermore, note that English-focused analyses based on our own machine translations of Russian speeches, speech titles, and related metadata within these Russian-language datasets are analytically distinct from analyses using the Russian government’s officially released English-language versions included in the English corpora. The latter may reflect not only translation but also selective editorial or framing decisions regarding what content is translated and how. Consequently, the appropriate choice of English-language text should be guided by a user’s specific research question. 

Researchers working with the Russian-language corpora should also be attentive to character encoding when reading files, as Cyrillic text may not render correctly under default settings. In \texttt{Python 3}~\cite{python3}, we recommend reading files explicitly as UTF-8 (e.g., \texttt{pandas.read\_csv(..., encoding="utf-8")} and, when needed, \texttt{encoding\_errors="strict"} to surface decoding problems rather than silently replacing characters); if decoding fails, users should avoid ad-hoc spreadsheet ``repairs'' and instead re-export or re-save the source file in UTF-8 and re-read it. In \texttt{R}, we recommend UTF-8-capable import routines such as \texttt{readr::read\_csv(..., locale = readr::locale(encoding = "UTF-8"))} or \texttt{data.table::fread(..., encoding = "UTF-8")}, and users should verify correct rendering by inspecting a few known Cyrillic tokens after import. 

Finally, opening the CSV files directly in spreadsheet software such as Microsoft Excel or Google Sheets is discouraged. Both tools impose per-cell character limits (Excel supports up to 32{,}767 characters per cell and Google Sheets effectively caps cells at 50{,}000 characters), and some speeches in our datasets contain text fields exceeding these thresholds; as a result, long texts may be truncated or omitted without obvious warning. These limitations are language-agnostic and can therefore affect both the Russian- and English-language CSVs whenever speech texts are very long. In addition, spreadsheet programs may apply automatic type inference and may mishandle Unicode during import/export workflows, increasing the risk of silent character corruption in Cyrillic fields if files are opened and re-saved. We therefore recommend accessing and processing the released CSV files using programmatic tools (e.g., Python or R) rather than spreadsheet editors, and preserving UTF-8 encoding throughout the workflow.

\section*{Code availability}

All code used for web scraping, data cleaning, translation, variable construction, topic modeling, and validation is publicly available in a GitHub repository (\texttt{https://github.com/bagozzib/Russian-Speech-Text-and-Images}). The repository provides a structured replication workflow, configuration files, and documentation describing how to reproduce the four final CSV archives (\texttt{kremlin\_english.csv}, \texttt{kremlin\_russian.csv}, \texttt{mid\_english.csv}, \texttt{mid\_russian.csv}), the four corresponding image archives, and the auxiliary long-format topic--probability tables and HTML topic summaries.

All analysis scripts were developed and executed in \texttt{Python 3} (any modern 3.x release should be sufficient; we recommend \texttt{Python} $\ge$3.10 for best compatibility with current NLP and scientific-computing packages) and/or \texttt{R} (any modern 4.x release). To facilitate reproducibility, the GitHub repository mentioned above provides explicit environment specifications (e.g., \texttt{requirements.txt} or equivalent) that pin package versions for the primary pipelines.

\section*{Data Availability}

All data described in this paper are publicly available via Harvard Dataverse \cite{BlionvaEtAlSGI0VK_2026}. The release is organized into four speech-level CSV tables (one per corpus), four corresponding image archives, and Auxiliary materials. Each component is described below.

\vspace{0.5em}
\noindent\textbf{CSV archives.}
Four analysis-ready, speech-level CSV tables (one per source-language corpus) that contain the full set of scraped metadata and the final modeling outputs used in the paper. The following CSV files can specifically be found within the kremlin\_mid\_en\_ru\_final\_csvs.zip on the abovementioned Dataverse page:

\begin{itemize}
  \item \texttt{kremlin\_english.csv}: Kremlin corpus in English (original English pages).
  \item \texttt{kremlin\_russian.csv}: Kremlin corpus in Russian, with the translated English text fields (e.g., \texttt{full\_text\_english}) used for topic modeling while preserving the original Russian content.
  \item \texttt{mid\_english.csv}: MID.ru corpus in English (original English pages).
  \item \texttt{mid\_russian.csv}: MID.ru corpus in Russian, with translated English text fields used for topic modeling while preserving the original Russian content.
\end{itemize}

\vspace{0.5em}
\noindent\textbf{Image archives.}
All scraped images referenced by the CSVs are distributed as separate zipped files corresponding to our four separate corpus-level archives (i.e., one per corpus):

\begin{itemize}
  \item \texttt{kremlin\_english\_images.zip}
  \item \texttt{kremlin\_russian\_images.zip}
  \item \texttt{mid\_english\_images.zip}
  \item \texttt{mid\_russian\_images.zip}
\end{itemize}

\noindent Within each corpus-specific archive, image files are stored in standard web formats (predominantly \texttt{.jpg}, with occasional \texttt{.png}). The \texttt{stored\_image\_filepaths} column in each CSV provides the authoritative linkage from speech records to image files: each cell contains a list of image paths that are \emph{relative to the corresponding corpus image root directory}. To locate an image, extract the relevant corpus archive and concatenate the corpus image root directory with the relative path listed in \texttt{stored\_image\_filepaths}. Image captions, when available, are provided in the parallel list-valued \texttt{image\_captions} field.

\vspace{0.5em}
\noindent\textbf{Auxiliary materials distributed with the data deposit.}
Alongside the four CSVs and four image archives, the Dataverse deposit includes a compact set of auxiliary resources to support inspection and reuse. The following file and folder names are each found within the kremlin\_mid\_en\_ru\_auxiliary\_files.zip file on the Harvard Dataverse page mentioned above:

\begin{itemize}
  \item \texttt{topic\_summaries\_html.zip}: HTML topic-summary files (one per learned topic per corpus) for qualitative inspection of keywords, representative speeches, and representative images.
\item \texttt{text\_and\_image\_topic\_probability\_files.zip}: long-format document--topic and image--topic probability tables exported from the modeling pipeline. The contents include:

\begin{itemize}
  \item \texttt{text\_probability\_files}
  \begin{itemize}
    \item \texttt{kremlin\_english\_text\_topic\_probs.csv}
    \item \texttt{kremlin\_russian\_text\_topic\_probs.csv}
    \item \texttt{mid\_english\_text\_topic\_probs.csv}
    \item \texttt{mid\_russian\_text\_topic\_probs.csv}
  \end{itemize}

  \item \texttt{image\_probability\_files}
  \begin{itemize}
    \item \texttt{kremlin\_english\_image\_topic\_probs.csv}
    \item \texttt{kremlin\_russian\_image\_topic\_probs.csv}
    \item \texttt{mid\_english\_image\_topic\_probs.csv}
    \item \texttt{mid\_russian\_image\_topic\_probs.csv}
  \end{itemize}
\end{itemize}
\end{itemize}

\noindent Together, these materials provide (i) analysis-ready speech-level tables, (ii) the complete set of linked image files, and (iii) optional topic-level and probability-level outputs that support robustness checks, alternative topic assignment strategies, and qualitative validation. These materials are also included on the project's Harvard Dataverse dataset page \cite{BlionvaEtAlSGI0VK_2026}.

\bibliography{russian_speech_dataset}

\section*{Acknowledgments} 

This work was supported in part by the National Science Foundation under Award No. 2417814, SCIPE: Building a Computational and Data-Intensive Research Workforce \& Network in the Mid-Atlantic Region (Strengthening the Cyberinfrastructure Professionals Ecosystem).

\section*{Author contributions statement}
B.B.\ and D.B.\ conceived the project. D.B.\, G.E.\, K.S.\, and R.E.\ implemented components of the study's webscrabing tasks. G.E.\, K.S.\, and R.E.\ implemented components of the study's topic modeling tasks and dataset extensions. B.B., D.B.\, M.R.\, handled components of topic labeling and validation. B.B.\, D.B.\, R.E.\, and S.C., helped to oversee project tasks, training, and coordination. All authors wrote and reviewed the manuscript.

\section*{Competing interests} The authors declare no competing interests.

\end{document}